\pdfoutput=1
\documentclass{article}
\usepackage[final,eandd]{neurips_2026}

\usepackage[utf8]{inputenc}
\usepackage[T1]{fontenc}
\usepackage{hyperref}
\usepackage{url}
\usepackage{booktabs}
\usepackage{amsfonts}
\usepackage{amsmath}
\usepackage{nicefrac}
\usepackage[expansion=false]{microtype}
\usepackage{graphicx}
\usepackage{multirow}
\usepackage{enumitem}
\usepackage{pifont}

\usepackage{tcolorbox}
\tcbuselibrary{breakable, skins}
\usepackage{caption}
\usepackage{wrapfig}

\newcommand{\cmark}{\ding{51}}
\newcommand{\xmark}{\ding{55}}

\title{OmniMemBench: Towards Scalable Evaluation \\ of Long-Term Omni-Modal Agent Memory}

\author{%
  \textbf{Junhan Shi}\textsuperscript{1,$\ddagger$}\quad
  \textbf{Qiyi Wang}\textsuperscript{3,2}\quad
  \textbf{Zhenning Shi}\textsuperscript{1}\quad
  \textbf{Xiongwei Wu}\textsuperscript{2}\\[2pt]
  \textbf{Ming Ma}\textsuperscript{5,2}\quad
  \textbf{Baoqi Pei}\textsuperscript{2}\quad
  \textbf{Yong Jiang}\textsuperscript{1,4}\quad
  \textbf{Qing Li}\textsuperscript{4}\\[2pt]
  \textbf{Weigao Sun}\textsuperscript{2,$\dagger$}\quad
  \textbf{Yiran Zhong}\textsuperscript{2,$\dagger$}\quad
  \textbf{Steven Hoi}\textsuperscript{2}\\[5pt]
  \textsuperscript{1}Tsinghua University\quad
  \textsuperscript{2}Tongyi Lab, Alibaba Group\\[2pt]
  \textsuperscript{3}Nanjing University\quad
  \textsuperscript{4}Peng Cheng Laboratory\\[2pt]
  \textsuperscript{5}School of Future Technology, University of Chinese Academy of Sciences
}

\begin{document}

\maketitle
\begingroup
\renewcommand{\thefootnote}{\fnsymbol{footnote}}
\footnotetext[2]{Corresponding authors.}
\footnotetext[3]{Work done during internship at Tongyi Lab, Alibaba Group.}
\endgroup

\begin{abstract}
Multimodal agents that interact with users over days or months must remember not only what was said, but also what was shown, heard, and played. Current memory benchmarks offer limited modality coverage and often rely on aggregate scores that conflate retrieval and answer-generation errors. We introduce OmniMemBench, a long-term memory benchmark of multi-session dialogues that interleave images, audio, and video, with 103 characters and 6,655 QAs across eight task types and four context tiers. An anti-leakage pipeline filters out questions that a text-only model can answer from the dialogue alone. Two LLM-judged retrieval metrics, Clue Coverage and Entry Relevance, score the retrieved content rather than entry IDs, so RAG and memory systems can be diagnosed on the same scale. Injecting unrelated distractor sessions scales the context from 128K to 1M tokens while keeping the evidence and answers fixed. We evaluate 13 pipeline configurations and seven full-context models with GLM-5.2 as the judge, and validate its answer and retrieval scores against independent human annotations. The strongest full-context models score highest overall, while simple retrieval remains better at single-hop fact lookup. Converting media to captions works better than indexing raw media in all four methods that support both, yet case studies show that captions can drop perceptual details, raw-media retrieval can miss the required entry, and the QA model can misread correctly retrieved media. Retrieving more entries improves QA, whereas a larger model for memory construction brings no consistent gain. OmniMemBench provides a common testbed for studying how multimodal agents retain, retrieve, and use past information.
The \href{https://github.com/jshmshijh/OmniMemBench}{evaluation code} and \href{https://huggingface.co/datasets/jshmsjh/OmniMemBench}{benchmark dataset} are publicly available.
\end{abstract}

\section{Introduction}
\label{sec:intro}

Multimodal agents operating in real-world settings must maintain memory across extended interactions. Users share photographs, voice messages, and video clips that may be referenced, compared, or updated in future conversations spanning days or months. Long-term memory lets an agent recall what was discussed, what was shown, and what has changed across these conversations.

Several lines of work address this problem: RAG-based methods~\citep{lewis2020rag} index conversation history for on-demand retrieval; cognitive-inspired systems~\begin{NoHyper}\citep{park2023generativeagents,packer2023memgpt}\end{NoHyper} compress and reorganize information into structured memory entries; memory frameworks~\citep{chhikara2025mem0,xu2025amem} offer plug-in long-term memory layers; and recent long-context models attempt to process entire interaction histories directly. However, few benchmarks evaluate these approaches on a common footing, especially when the interactions are multimodal. Existing benchmarks leave three gaps (Table~\ref{tab:comparison}):

\textbf{Gap 1: Incomplete modality coverage without verified media dependence.}
None of these benchmarks covers image, audio, and video together: LongMemEval~\citep{wu2024longmemeval} and BEAM~\citep{tavakoli2025beam} are text-only; Mem-Gallery~\citep{bei2026memgallery} adds images but lacks audio and video; ATM-Bench~\citep{mei2026atmbench} covers images, videos, and emails but omits audio and operates outside conversational settings. In addition, few verify that the questions cannot be answered from the surrounding text alone.

\textbf{Gap 2: No cross-paradigm failure diagnosis.}
Memory benchmarks mainly score final answers. When they also evaluate retrieval, they match the retrieved items against annotated evidence IDs~\citep{maharana2024locomo,bei2026memgallery,mei2026atmbench}. This works when retrieved items keep the IDs of the original dialogue, as RAG chunks do, but not when memory methods rewrite or merge information into new entries. Retrieval quality is therefore hard to compare across paradigms, and a low QA score does not reveal whether a system failed at storage, retrieval, or reasoning.

\textbf{Gap 3: Limited scalability evaluation, especially for multimodal settings.}
Real agent interactions grow continuously over time. While LongMemEval~\citep{wu2024longmemeval} and BEAM~\citep{tavakoli2025beam} support context scaling in text-only settings, multimodal memory benchmarks offer limited controlled scalability evaluation. Holding evidence fixed while adding distractors tests how a system handles a growing history, beyond its performance at a single context length.

\paragraph{Our Approach.}
OmniMemBench addresses each of these gaps. For \textbf{Gap~1}, it covers image (49.9\%), audio (39.6\%), and video (10.5\%), and an anti-leakage pipeline filters out questions whose answers can be largely inferred from the text alone. For \textbf{Gap~2}, two LLM-judged metrics score the retrieved content instead of entry IDs: Clue Coverage measures whether the evidence needed for an answer appears among the retrieved entries, and Entry Relevance measures how much of the retrieved content is relevant to the question. Both apply equally to RAG and memory methods, and together with QA scores and case studies they help identify where information is lost along the pipeline. For \textbf{Gap~3}, content-isolated distractor injection produces four context tiers (128K--1M tokens), with evidence content held constant across tiers. The benchmark comprises 103 characters with 2,719 QAs at the base 128K tier, scaling to 6,655 QAs across all four tiers, constructed from diverse source datasets through a hierarchical pipeline. The same data can also be given directly to long-context multimodal models, so all paradigms are compared on identical inputs.

\paragraph{Key Findings.}
We evaluate 13 pipeline configurations of long-context, RAG, and memory methods, together with seven full-context models. Our main findings are as follows:
\begin{itemize}[leftmargin=*,nosep]
    \item \textbf{The strongest full-context models lead overall, while retrieval wins on precise lookups.} On the Omni track, Gemini-3.1-Pro scores 0.750 overall, ahead of the RAG and memory baselines, but NaiveRAG scores higher on single-hop questions (0.980 vs.\ 0.816; Table~\ref{tab:omni_track}).
    \item \textbf{Simple retrieval outperforms the tested memory systems.} In the main comparison, NaiveRAG reaches 0.606, compared with 0.520 for the strongest memory method, AUGUSTUS. The gap is largest on multi-hop and knowledge-update questions.
    \item \textbf{Captions work better than raw media in current pipelines, but both routes have failure modes.} Caption mode scores higher than raw mode in all four methods that support both. Case studies illustrate three failure types: captions drop perceptual details, raw-media retrieval returns topically similar but wrong entries, and the QA model can misread media that was correctly retrieved.
    \item \textbf{Retrieving more helps, while a larger store model does not consistently help.} QA improves as the number of retrieved entries grows from 5 to 20, and Clue Coverage is positively associated with QA across methods. Replacing the 35B store backbone with a 122B model changes overall scores by at most 0.016 in the tested configurations.
\end{itemize}

\paragraph{Contributions.}
\textbf{(1)}~We introduce OmniMemBench, which to our knowledge is the first omni-modal memory benchmark in multi-session conversational settings. It covers image, audio, and video, applies systematic anti-leakage checks, and includes eight complementary task types.
\textbf{(2)}~We propose two content-based retrieval metrics, Clue Coverage and Entry Relevance, validate them against independent human annotations, and compare them with ID-based retrieval metrics.
\textbf{(3)}~We provide a scalable evaluation framework that tests memory systems and long-context models on identical data across four context tiers (128K--1M).
\textbf{(4)}~We empirically compare memory, RAG, and full-context approaches, analyzing media representation, retrieval depth, backbone choice, and context growth.

\section{Related Work}
\label{sec:related}

\subsection{Long-term Memory Benchmarks}

Several benchmarks evaluate conversational memory, but none simultaneously provides omni-modal coverage, evidence traceability, context scaling, and systematic checks of multimodal dependence (Table~\ref{tab:comparison}).

LoCoMo~\citep{maharana2024locomo} tests five reasoning types over multi-session dialogues with images, but its questions can largely be answered without the images~\citep{bei2026memgallery}.
LongMemEval~\citep{wu2024longmemeval} and BEAM~\citep{tavakoli2025beam} introduce evidence tracing and context scaling for text-only memory, with BEAM scaling up to 10M tokens across ten task types.
Mem-Gallery~\citep{bei2026memgallery} constructs questions that require visual evidence, but covers only images and does not scale context length.
ATM-Bench~\citep{mei2026atmbench} spans images, videos, and emails over $\sim$4 years of personal data, but omits audio, operates outside conversational settings, and does not verify multimodal irreplaceability.
MemoryAgentBench~\citep{hu2025memoryagentbench} evaluates memory competencies through incremental interactions, but is text-only.
Other recent benchmarks target different aspects: MemBench~\citep{tan2025membench} introduces efficiency and capacity metrics, Mem2ActBench~\citep{shen2026mem2actbench} evaluates proactive memory-to-action grounding, and M3-Bench~\citep{long2025m3agent} tests memory-based reasoning over audio-visual streams; these are either text-only or outside multi-session conversational settings.
Unlike single-context multimodal benchmarks~\citep{fu2024videomme,li2024omnibench,yue2024mmmu} that evaluate perception and reasoning within the current input, OmniMemBench targets \emph{memory}: storing, retrieving, and reasoning over multimodal information encountered across temporally separated sessions.

\begin{table}[t]
\centering
\caption{Comparison with existing memory benchmarks. \textbf{Modalities}: T = Text, I = Image, A = Audio, V = Video. \textbf{MM Irr.}: systematic checks that perceptually grounded QAs require multimodal input (``---'' = text-only, not applicable). OmniMemBench reports the 128K base tier and all four tiers (128K--1M) separately.}
\label{tab:comparison}
\small
\setlength{\tabcolsep}{3.5pt}
\begin{tabular}{@{}lcrccccc@{}}
\toprule
Benchmark & Modalities & \#QAs & \#Types & Sess. & Scaling & Evidence & MM Irr. \\
\midrule
MemoryAgentBench~\citep{hu2025memoryagentbench} & T & 2,071 & 4 & \cmark & \xmark & \xmark & --- \\
LongMemEval~\citep{wu2024longmemeval} & T & 500 & 7 & \cmark & \cmark & \cmark & --- \\
BEAM~\citep{tavakoli2025beam} & T & 2,000 & 10 & \cmark & \cmark & \cmark & --- \\
ATM-Bench~\citep{mei2026atmbench} & T+I+V & 1,038 & --- & \xmark & \xmark & \cmark & \xmark \\
LoCoMo~\citep{maharana2024locomo} & T+I & 7,512 & 5 & \cmark & \xmark & \cmark & \xmark \\
Mem-Gallery~\citep{bei2026memgallery} & T+I & 1,711 & 9 & \cmark & \xmark & \cmark & \cmark \\
\midrule
\textbf{Ours} (128K) & \textbf{T+I+A+V} & 2,719 & 8 & \cmark & \cmark & \cmark & \cmark \\
\textbf{Ours} (all tiers) & \textbf{T+I+A+V} & 6,655 & 8 & \cmark & \cmark & \cmark & \cmark \\
\bottomrule
\end{tabular}
\end{table}

\subsection{Agent Memory Methods}

Memory methods span several paradigms: \emph{long-context} approaches feed entire histories to extended-window models~\citep{team2024gemini}; \emph{RAG-based} methods index and retrieve conversation chunks~\citep{lewis2020rag}, with multimodal extensions via modality-aware routing~\citep{yeo2025universalrag}; \emph{structured memory} methods organize information into semantic tags~\citep{jain2025augustus}, entity-centric graphs~\citep{fisher2025neuralgraph}, or temporal knowledge graphs~\citep{rasmussen2025zep}; and \emph{cognitive-inspired} methods implement observation--reflection loops~\citep{park2023generativeagents}, OS-style memory management~\citep{packer2023memgpt}, or verbal self-critique~\citep{shinn2023reflexion}. Associative retrieval frameworks such as HippoRAG~\citep{gutierrez2024hipporag} use knowledge graphs with personalized PageRank for multi-hop memory recall. Memory frameworks such as Mem0~\citep{chhikara2025mem0} and A-Mem~\citep{xu2025amem} offer plug-in memory layers.
These methods are usually evaluated on different data with different metrics, almost always in text-only settings. OmniMemBench provides a common multimodal testbed for these paradigms.

\section{OmniMemBench}
\label{sec:benchmark}

\subsection{Overview}
\label{sec:overview}

OmniMemBench measures how well a system answers questions that depend on earlier sessions, which requires retaining information, reasoning about time, and tracking how facts change. RAG and structured memory are treated as two ways of providing this capability. QA scores reflect the whole pipeline, including perception and reasoning, while Clue Coverage and Entry Relevance (CC/ER; \S\ref{sec:evaluation}) evaluate the retrieved content directly.

\paragraph{Benchmark at a Glance.}
OmniMemBench comprises 103 characters, each with a distinct persona and a complete conversation history averaging 39.9 sessions and 400.0 dialogue turns. Eight task types (Table~\ref{tab:task_types}) span from precise extraction to global synthesis. Four context tiers (128K, 256K, 512K, 1M tokens) are constructed via distractor session injection: the 128K tier covers all 103 characters (2,719 QAs), while the 256K--1M tiers use a 51-character subset (1,312 QAs each) whose type and modality distributions closely match the full set, yielding 6,655 QAs in total. Figure~\ref{fig:data_structure} illustrates the data structure and task types.

\subsection{Construction Pipeline}
\label{sec:construction}

The benchmark is constructed through a hierarchical pipeline with six stages (Figure~\ref{fig:pipeline}). We highlight key design decisions below; full details are in Appendix~\ref{app:construction}.

\begin{figure}[t]
\centering
\includegraphics[width=0.94\linewidth]{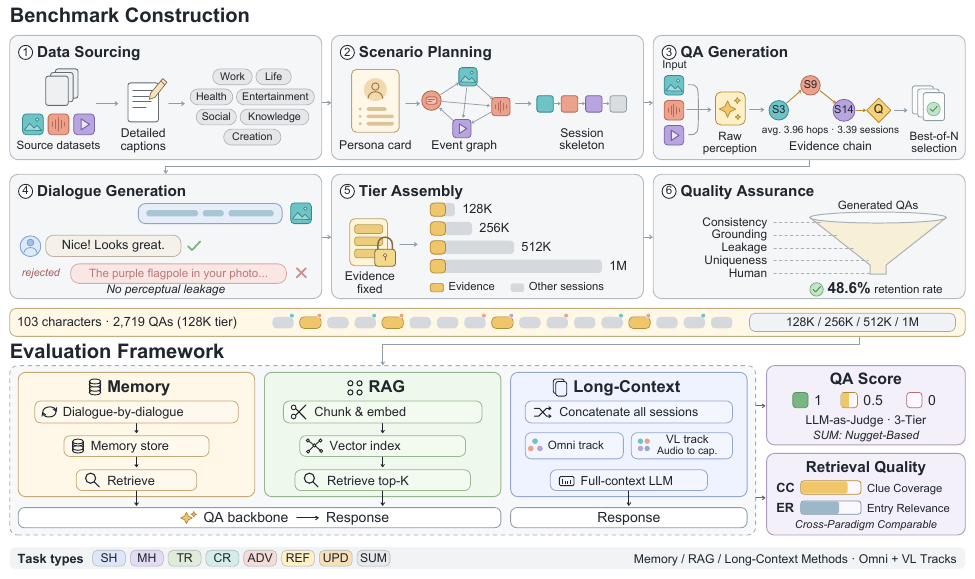}
\caption{OmniMemBench construction pipeline and evaluation framework. \emph{Top}: Six-stage construction from data sourcing to quality assurance. \emph{Bottom}: Unified evaluation framework for memory, RAG, and long-context paradigms with QA scoring (three levels for non-SUM tasks; nugget coverage for SUM) and cross-paradigm retrieval quality metrics (Clue Coverage and Entry \mbox{Relevance}).}
\label{fig:pipeline}
\end{figure}

\subsubsection{Data Sourcing \& Scenario Planning}

We source multimodal content from 10 publicly available datasets spanning image~\citep{plummer2015flickr30k,yue2024mmmupro}, audio~\citep{sakshi2024mmau}, and video~\citep{li2024mvbench} domains, as well as cross-modal benchmarks~\citep{li2024omnibench}. We use real media rather than generated media because details such as the visual content of a photograph, the timbre of an audio clip, or the motion in a video are hard to reproduce faithfully with text-conditioned generation. For each character, we generate a persona profile, an event graph capturing life events and their temporal/causal relationships, and a session skeleton that schedules when each multimodal file naturally enters the conversation.

\subsubsection{QA Generation}

The generation model receives the raw media files and their captions and constructs evidence chains that follow the event graph. Chains typically span multiple sessions and modalities, so answering requires cross-session reasoning grounded in multimodal content. We generate multiple candidates per question and select the version that maximizes difficulty while maintaining evidence closure---every hop in the chain must be verifiable from the conversation history.

\subsubsection{Dialogue Generation \& Anti-Leakage}
\label{sec:anti_leakage}

Sessions are generated as natural multi-turn dialogues, with evidence embedded in the conversational flow. The central design constraint is \emph{anti-leakage}: dialogue text must not reveal perceptual information from multimodal files. For example, when a user shares an image, the assistant must not describe its visual content in the reply. Without this constraint, a text-only system could answer multimodal questions by extracting leaked descriptions---a failure mode that recent work has found to be common in multimodal benchmarks, where models achieve high scores from textual cues without any image input~\citep{asadi2026mirage}.

We enforce anti-leakage by separating what a media file contains from how the dialogue refers to it. Each multimodal file is associated with: (1)~a \emph{caption}, a full perceptual description that the generation pipeline uses to plan dialogues and questions but never writes into the dialogue; (2)~a \emph{context hint}, a non-perceptual cue that lets the dialogue and the question refer to the file without describing it (e.g., ``check out this photo from the hiking trip''); and (3)~a \emph{question} about perceptual details recorded in the caption but absent from the dialogue, so that answering it requires the media. SUM questions do not follow this per-file scheme: they are written after the dialogues are complete, draw on many sessions, and are scored against atomic reference nuggets, so part of their content can be recovered from the dialogue text. Appendix~\ref{app:multimodel_leakage} tests how well the mechanism works by having four text-only models answer the questions with all media removed, and reports SUM separately.

\subsubsection{Context Scaling via Distractor Injection}

We scale context from 128K to 1M tokens by injecting distractor sessions around fixed evidence sessions. Distractors retain the character's persona and speaking style, while the character's event graph and questions are given to the generator as forbidden content, constraining distractors to unrelated events that neither supply answer evidence nor contradict established facts. This holds the intended evidence and answers constant as context grows. The token share of evidence-bearing dialogues falls from 57.7\% to 8.9\%. Appendix~\ref{app:tier_stats} provides tier statistics.

\subsubsection{Quality Assurance}

We apply multi-stage auditing: (1) consistency check verifies that the evidence chain logically supports the reference answer and that reasoning steps require no information outside the evidence; (2) grounding check confirms that every planned evidence hop is actually embedded in the generated dialogue and that the reference answer is not directly stated in dialogue text; (3) leakage filter tests each QA with a text-only model on media-removed dialogue, removing answerable candidates to reduce textual shortcuts on perceptually grounded tasks; (4) uniqueness check verifies that the answer is deterministic given the evidence, filtering out QAs with multiple plausible answers; and (5) human review by the author team for clarity, factual correctness, and edge cases that automated stages may miss. From an initial pool of ${\sim}$5,600 candidates from the 128K tier, 2,719 QAs survive all stages (48.6\% retention rate). Full details are in Appendix \ref{app:qa_pipeline}.

\subsection{Evaluation Protocol}
\label{sec:evaluation}

\paragraph{QA Evaluation.}
We use GLM-5.2 as the LLM judge~\citep{zheng2023judging} at temperature 0. Seven non-SUM types receive 1 (correct), 0.5 (partially correct), or 0 (incorrect). SUM uses nugget-based evaluation~\citep{voorhees2004trec}: each question has fixed reference nuggets shared across systems, and its score is the fraction covered by the answer. Overall is the mean over valid questions, including SUM. Prompts and aggregation details are in Appendix~\ref{app:eval_details}.

\paragraph{Retrieval Quality Evaluation.}
To enable cross-paradigm diagnosis, we propose two LLM-judged retrieval metrics that operate on \emph{content} rather than entry IDs, making them applicable to both RAG (with retrievable chunks) and memory systems (with restructured entries):

\begin{itemize}[leftmargin=*,nosep]
    \item \textbf{Clue Coverage (CC)}: For each ground-truth clue required to answer the question, the judge assesses whether it is covered by any retrieved entry. CC measures recall at the clue level.
    \item \textbf{Entry Relevance (ER)}: For each retrieved entry, the judge assesses whether it is relevant to the question. ER measures precision at the entry level.
\end{itemize}

\paragraph{Metric Validation.}
Two non-author annotators independently scored 650 answer items: human--human weighted agreement is $\kappa_w{=}0.889$, and GLM--human agreement is $\kappa_w{=}0.902$. Separate clue and entry annotations yield GLM--human $\kappa{=}0.753$ for CC and $0.787$ for ER. Agreement with ID-based IR metrics varies by evaluation set (Appendix~\ref{app:metric_validation}), so CC/ER complement these metrics rather than replace them. Full human-validation results are in Appendix~\ref{app:human_validation}.

\section{Experiments}
\label{sec:experiments}

\subsection{Experimental Setup}
\label{sec:setup}

\paragraph{Evaluated Methods.}
We evaluate nine methods spanning three paradigms: \textbf{Long-context} (SlidingWindow), \textbf{RAG} (NaiveRAG, MultimodalRAG, UniversalRAG~\citep{yeo2025universalrag}), and \textbf{Memory} (GenerativeAgents~\citep{park2023generativeagents}, NeuralGraph~\citep{fisher2025neuralgraph}, AUGUSTUS~\citep{jain2025augustus}, AMem~\citep{xu2025amem}, Mem0~\citep{chhikara2025mem0}).
The four methods with native multimodal support (MultimodalRAG, UniversalRAG, NeuralGraph, AUGUSTUS) are each tested in \emph{caption} (media converted to text) and \emph{raw} (media indexed with multimodal embeddings) modes, yielding $9 + 4 = 13$ configurations.
Full method descriptions are in Appendix~\ref{app:methods}.

\paragraph{Default Configuration.}
Unless stated otherwise, methods that use an LLM during memorization use Qwen3.5-35B-A3B~\citep{qwen3} as the store backbone, and all methods answer with Gemini-3-Flash as the QA backbone. Each query retrieves $K{=}10$ entries, experiments run on the 128K tier (103 characters, 2,719 QAs), and GLM-5.2 is the judge. Comparisons across backbones and context tiers use, for each method, the questions with valid results in every compared setting (Appendix~\ref{app:aggregation}).

\paragraph{Research Questions.}
We organize the experiments around five research questions:
\begin{itemize}[leftmargin=*,nosep]
    \item \textbf{RQ1:} How do long-context, RAG, and memory methods compare on multimodal memory tasks?
    \item \textbf{RQ2:} How does the way media is processed (caption vs.\ raw) affect multimodal methods?
    \item \textbf{RQ3:} How do retrieval quantity and quality affect QA performance?
    \item \textbf{RQ4:} How robust are methods as the context grows from 128K to 1M tokens?
    \item \textbf{RQ5:} How do methods trade off accuracy against computational cost?
\end{itemize}

\subsection{Main Results (RQ1 \& RQ2)}
\label{sec:main_results}

Table~\ref{tab:main_results} presents results across 13 configurations and eight task types.

\begin{table*}[t]
\centering
\caption{Main results under GLM-5.2 at 128K (103 characters; 2,719 benchmark QAs). $^{\mathrm{c}}$: caption; $^{\mathrm{r}}$: raw. Overall includes SUM. Best displayed scores are bold. Aggregation details are in Appendix~\ref{app:aggregation}.}
\label{tab:main_results}
\small
\setlength{\tabcolsep}{2.8pt}
\begin{tabular*}{\linewidth}{@{\extracolsep{\fill}}llrrrrrrrrr@{}}
\toprule
Method & Paradigm & SH & MH & TR & CR & ADV & REF & UPD & SUM & Overall \\
\midrule
NaiveRAG$^{\mathrm{c}}$ & RAG & \textbf{.976} & \textbf{.566} & .536 & .670 & \textbf{.939} & .488 & .419 & .210 & \textbf{.606} \\
MM-RAG$^{\mathrm{c}}$ & RAG & .971 & .559 & \textbf{.548} & \textbf{.675} & .937 & .439 & .428 & .202 & .603 \\
UniversalRAG$^{\mathrm{c}}$ & RAG & .944 & .289 & .467 & .450 & .867 & .794 & \textbf{.479} & \textbf{.211} & .528 \\
AUGUSTUS$^{\mathrm{c}}$ & Memory & .959 & .336 & .519 & .587 & .922 & .536 & .254 & .162 & .520 \\
MM-RAG$^{\mathrm{r}}$ & RAG & .706 & .309 & .481 & .576 & .882 & .530 & .246 & .192 & .476 \\
UniversalRAG$^{\mathrm{r}}$ & RAG & .722 & .243 & .437 & .438 & .808 & .797 & .370 & .201 & .466 \\
AUGUSTUS$^{\mathrm{r}}$ & Memory & .723 & .294 & .468 & .539 & .852 & .620 & .228 & .153 & .464 \\
NeuralGraph$^{\mathrm{c}}$ & Memory & .946 & .230 & .396 & .481 & .876 & .642 & .153 & .126 & .452 \\
AMem$^{\mathrm{c}}$ & Memory & .751 & .225 & .404 & .550 & .886 & .539 & .169 & .158 & .438 \\
NeuralGraph$^{\mathrm{r}}$ & Memory & .649 & .163 & .388 & .418 & .767 & .700 & .088 & .123 & .376 \\
Mem0$^{\mathrm{c}}$ & Memory & .175 & .098 & .414 & .345 & .573 & .767 & .062 & .071 & .273 \\
Gen.Agents$^{\mathrm{c}}$ & Memory & .178 & .019 & .181 & .269 & .863 & .930 & .049 & .041 & .259 \\
SlidingWin$^{\mathrm{c}}$ & Long-ctx & .078 & .010 & .128 & .176 & .761 & \textbf{.958} & .015 & .025 & .210 \\
\bottomrule
\end{tabular*}
\end{table*}

\paragraph{The best RAG method outperforms the best memory method overall (RQ1).}
NaiveRAG (0.606) exceeds the strongest memory method, AUGUSTUS-caption (0.520), by 0.086, and MultimodalRAG-caption is close behind at 0.603.
The two leading RAG methods score above 0.97 on single-hop (\textsc{SH}) questions, and the best memory system comes close (0.959 for AUGUSTUS-caption), but these two methods are much stronger on multi-hop (\textsc{MH}) and knowledge-update (\textsc{UPD}) questions: NaiveRAG reaches 0.566 and 0.419 on these types, compared with 0.336 and 0.254 for AUGUSTUS-caption.
Task-level scores also show why a single task type can mislead. SlidingWindow, which keeps only the most recent 32K tokens, scores 0.958 on answer refusal (\textsc{REF}) but only 0.078 on \textsc{SH} and 0.015 on \textsc{UPD}; since refusing is the correct response to \textsc{REF} questions, a system that retains little can still score well on them. Several configurations exceed 0.9 on adversarial (\textsc{ADV}) questions, whereas summarization (\textsc{SUM}) remains hard: no pipeline covers more than 0.211 of the reference nuggets.

\paragraph{Caption mode consistently outperforms raw mode (RQ2).}
Converting media to captions before storage yields higher overall scores than indexing the raw media in all four methods that support both modes: AUGUSTUS +0.057, MultimodalRAG +0.127, UniversalRAG +0.063, and NeuralGraph +0.076.
The largest gaps appear on \textsc{SH} questions (0.222 to 0.297), which require finding one specific piece of evidence, suggesting that raw-media retrieval struggles with precise matching. Raw mode scores higher only on \textsc{REF}. Section~\ref{sec:discussion} examines concrete failure cases.

\subsection{Retrieval Analysis (RQ3)}
\label{sec:retrieval}

\begin{figure}[t]
\centering
\includegraphics[width=\linewidth]{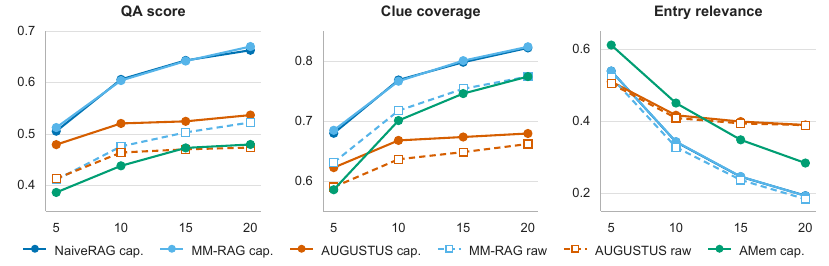}
\caption{Retrieval depth for six configurations. QA uses each method's common questions across $K{=}5,10,15,20$; CC/ER use valid retrieval judgments at each $K$. Larger retrieved sets improve QA and coverage while reducing entry relevance.}
\label{fig:k_sensitivity}
\end{figure}

\paragraph{Effect of Retrieval $K$.}
Retrieving more entries helps every tested configuration: QA rises monotonically from $K{=}5$ to $20$ for all six configurations, each measured on its common question set (Figure~\ref{fig:k_sensitivity}). The two caption-mode RAG methods benefit most, with NaiveRAG and MultimodalRAG-caption each gaining about 0.157, compared with 0.057 for AUGUSTUS-caption. Larger retrieved sets raise CC but lower ER: more of the needed evidence is retrieved, along with more irrelevant entries. Most of the QA gain occurs between $K{=}5$ and $10$, so we use $K{=}10$ as the default; all four settings are reported in Appendix~\ref{app:k_sensitivity}.

\begin{table}[t]
\centering
\caption{Retrieval quality and QA at $K{=}10$. CC measures clue coverage; ER measures entry relevance. Neither alone determines answer quality.}
\label{tab:cc_qa}
\small
\setlength{\tabcolsep}{3.5pt}
\begin{tabular*}{\linewidth}{@{\extracolsep{\fill}}llrrr@{}}
\toprule
Method & Paradigm & CC & ER & QA \\
\midrule
NaiveRAG$^{\mathrm{c}}$ & RAG & .769 & .342 & .606 \\
MM-RAG$^{\mathrm{c}}$ & RAG & .767 & .343 & .603 \\
AUGUSTUS$^{\mathrm{c}}$ & Memory & .668 & .417 & .520 \\
MM-RAG$^{\mathrm{r}}$ & RAG & .718 & .328 & .476 \\
AUGUSTUS$^{\mathrm{r}}$ & Memory & .637 & .409 & .464 \\
AMem$^{\mathrm{c}}$ & Memory & .701 & .451 & .438 \\
Mem0$^{\mathrm{c}}$ & Memory & .517 & .303 & .273 \\
\bottomrule
\end{tabular*}
\end{table}

\paragraph{Coverage and Relevance Provide Different Diagnostics.}
Table~\ref{tab:cc_qa} compares retrieval quality with QA at $K{=}10$. Across the seven configurations, CC is positively associated with QA (Spearman $\rho{=}0.821$), whereas ER is not ($\rho{=}0.036$), so coverage tracks answer quality more closely than relevance does. AMem, for example, has higher ER than NaiveRAG (0.451 vs.\ 0.342) but much lower QA (0.438 vs.\ 0.606). Coverage alone does not determine QA either: AMem's coverage is 0.068 below NaiveRAG's, yet its QA is 0.168 lower, and coverage varies widely among memory methods, from 0.517 for Mem0 to 0.701 for AMem. Answer quality also depends on whether the retrieved entries keep the details a question needs and on how the QA model uses them, so aggregate scores alone cannot show where a given pipeline fails. The case studies in \S\ref{sec:discussion} examine individual failures.

\subsection{Context Scaling (RQ4)}
\label{sec:scaling}

\begin{figure}[t]
\centering
\begin{minipage}[t]{0.44\linewidth}
    \centering
    \includegraphics[width=\linewidth]{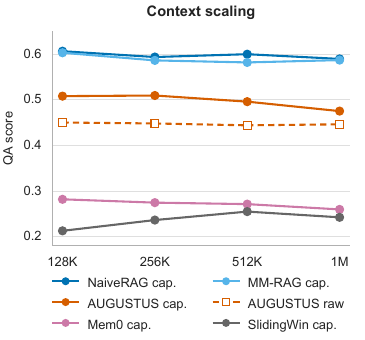}
\end{minipage}
\hfill
\begin{minipage}[t]{0.54\linewidth}
    \centering
    \includegraphics[width=\linewidth]{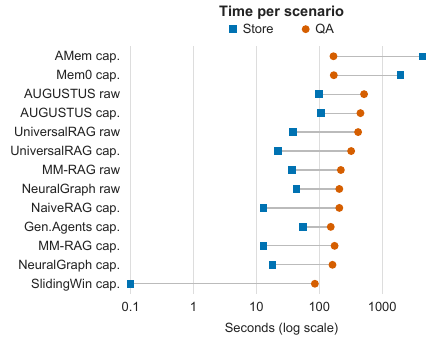}
\end{minipage}
\caption{\textbf{Left}: Context scaling on the 51-character subset, using each method's common valid \mbox{questions} across all four tiers. \textbf{Right}: Store and QA time per scenario at 128K (log scale). NaiveRAG combines the highest pipeline score (0.606) with 221s per scenario.}
\label{fig:scaling_efficiency}
\end{figure}

Figure~\ref{fig:scaling_efficiency} (left) tracks six configurations from 128K to 1M tokens, each on its common question set. The two caption-mode RAG methods keep the highest scores at every tier and lose only 0.017 (NaiveRAG) and 0.016 (MultimodalRAG) from 128K to 1M. AUGUSTUS-caption drops by 0.033 and Mem0 by 0.022, while AUGUSTUS-raw is nearly flat ($-$0.004) and SlidingWindow even rises by 0.030. A flat curve does not make a method strong, however: AUGUSTUS-raw and SlidingWindow both stay well below the RAG methods at every tier.

\subsection{Efficiency (RQ5)}
\label{sec:efficiency}

Figure~\ref{fig:scaling_efficiency} (right) shows the wall-clock time per scenario, split into memorization (store) and question answering.
NaiveRAG obtains the highest pipeline score (0.606) in 221s per scenario. AMem takes 4,548s, of which 4,380s is spent on memorization, and Mem0 spends 1,948s on memorization; both then answer the questions in about 170s. Higher cost therefore does not bring higher accuracy here: the two most expensive methods both score below NaiveRAG.
The full timing breakdown and per-tier store times are in Appendix~\ref{app:efficiency}.

\subsection{Discussion: Where Multimodal Information Is Lost}
\label{sec:discussion}

The quantitative results and the case studies in Appendix~\ref{app:cases} point to several places where the tested pipelines lose multimodal information.

\paragraph{Caption Route: Memorization-Time Information Loss.}
A retrieved memory is useful only if it still contains the detail that a question asks about. Captions turn media into text before storage, and they may keep a coarse description while dropping the detail that a later question needs. In Case~1, the stored audio captions describe the music as blues-style, but they omit the reggae-fusion rhythm and the male vocalist required by the reference answer. Because memorization happens before any question is asked, a system must decide what to keep without knowing which details will matter later. A larger store backbone does not consistently help in the tested configurations (Appendix~\ref{app:backbone}).

\paragraph{Raw Route: Retrieval-Time Alignment Failure.}
Keeping the original media avoids this loss, but retrieval then depends on matching a text query against media embeddings. Raw mode scores lower than caption mode in all four paired methods, by 0.081 on average. In Case~2, raw-mode retrieval returns entries about reverb rules, such as rejecting heavy reverb, and the required audio entry falls out of the top-$K$, whereas caption-based retrieval finds it. Even correctly retrieved media can be misread: in Case~3, the required image reaches the QA model, which describes the purple flagpole as grey. The three cases thus fail at different stages: memorization, retrieval, and answering.

\section{OmniMemBench as a Long-Context Benchmark}
\label{sec:longcontext}

OmniMemBench can also be used as an omni-modal long-context benchmark: all sessions are concatenated into the model input, so long-context models and memory pipelines are compared on the same data. We evaluate Gemini-3.1-Pro, Gemini-3-Flash~\citep{gemini31pro}, Qwen3.5-Omni-Plus, and Qwen3.5-Omni-Flash~\citep{qwen35omni} on the 128K tier (1,453 QAs).\footnote{The Gemini API limits the number of video files per request. We evaluate all characters satisfying this constraint.} The \emph{Omni track} preserves all original media; the \emph{VL track} converts audio to captions for models without native audio support.

\begin{table}[t]
\centering
\caption{Omni track at 128K (57 characters; 1,453 benchmark QAs). Full-context models receive original media; RAG/memory baselines use captions. GLM-5.2 judges all rows. Best displayed scores are bold.}
\label{tab:omni_track}
\small
\setlength{\tabcolsep}{2.0pt}
\begin{tabular*}{\linewidth}{@{\extracolsep{\fill}}llrrrrrrrrr@{}}
\toprule
Model / method & Paradigm & SH & MH & TR & CR & ADV & REF & UPD & SUM & Overall \\
\midrule
Gemini-3.1-Pro & Long-ctx & .816 & \textbf{.767} & \textbf{.753} & \textbf{.828} & .907 & .552 & \textbf{.607} & \textbf{.642} & \textbf{.750} \\
Gemini-3-Flash & Long-ctx & .784 & .721 & .670 & .796 & .902 & .435 & .555 & .560 & .698 \\
Qwen3.5-Omni-Plus & Long-ctx & .698 & .329 & .469 & .611 & .644 & .342 & .229 & .506 & .473 \\
Qwen3.5-Omni-Flash & Long-ctx & .555 & .244 & .469 & .361 & .850 & .312 & .137 & .317 & .400 \\
NaiveRAG$^{\mathrm{c}}$ & RAG & \textbf{.980} & .576 & .523 & .680 & \textbf{.945} & .481 & .310 & .205 & .594 \\
AUGUSTUS$^{\mathrm{c}}$ & Memory & .968 & .344 & .513 & .553 & .934 & \textbf{.597} & .187 & .152 & .511 \\
\bottomrule
\end{tabular*}
\end{table}

\paragraph{The strongest full-context models lead overall.}
Gemini-3.1-Pro scores 0.750, ahead of NaiveRAG (0.594) and AUGUSTUS (0.511) on this track. Its advantage over NaiveRAG is largest on SUM (0.642 vs.\ 0.205) and UPD (0.607 vs.\ 0.310), both of which draw on information spread across many sessions. Full-context access alone does not guarantee higher scores: both Qwen Omni models trail NaiveRAG overall.

\paragraph{RAG/Memory excel at precise retrieval.}
On SH, NaiveRAG (0.980) and AUGUSTUS (0.968) surpass Gemini-3.1-Pro (0.816). Targeted search thus remains effective for factual lookup even where the strongest full-context models lead overall, which suggests combining selective retrieval with broad context access.

\paragraph{VL track.}
The VL track replaces audio with captions for models without native audio support; results for six models, including Kimi-K2.5~\citep{kimiteam2026kimik25}, are in Appendix~\ref{app:vl_track}. Besides this change in input, the VL track scores non-SUM answers against expanded reference answers, so its scores are not directly comparable with those of the Omni track.

\section{Conclusion}
\label{sec:conclusion}

We introduced OmniMemBench, an omni-modal memory benchmark with evidence-traceable tasks and four context tiers (128K--1M). Evaluation of 13 pipeline configurations and seven full-context models shows complementary strengths: the strongest full-context models lead overall, while targeted retrieval excels at single-hop extraction. Content-based retrieval diagnostics and case studies expose failures in retaining, retrieving, and interpreting multimodal information. Future systems may benefit from compact indices that keep access to the original media, together with stronger answer generation.

\paragraph{Limitations.}
Our dialogues are synthetic: this allows controlled evaluation, but it remains to be tested how well the results carry over to real long-term interactions. Scores reflect end-to-end performance on long-term memory tasks, which combine retaining, retrieving, and reasoning over past multimodal information, rather than an isolated measure of memory ability. Gemini models generate the dialogues and questions, and the Gemini systems score highest; a subset regenerated with \mbox{GPT-5.5} preserves the paradigm ranking (Appendix~\ref{app:generator_robustness}), but this does not fully rule out model-family bias. Human validation of the judge covers sampled outputs rather than all responses (Appendix~\ref{app:human_validation}). Contexts stop at 1M tokens, video accounts for only 10.5\% of the media, and the Omni track includes only models that natively accept audio, image, and video at 128K tokens. Finally, the leakage filter uses a single text-only model and may remove some valid questions whose answers happen to be guessable from text; text-only models from four families also score low on the core task types once the media are removed (Appendix~\ref{app:multimodel_leakage}).

\subsection*{AI Use Statement}
\label{sec:aiuse}
LLMs serve three roles in this work. In construction, Gemini models generate persona profiles, event graphs, dialogues, and QA candidates under the structural constraints described in \S\ref{sec:construction}; all outputs undergo the five-stage audit pipeline (\S\ref{app:qa_pipeline}) with final review by the author team. GLM-5.2 serves as the answer and retrieval judge (\S\ref{sec:evaluation}), and Gemini-3-Flash / Gemini-3.1-Pro serve as QA backbones (\S\ref{sec:setup}). Additional generators and judges are identified in Appendices~\ref{app:generator_robustness} and~\ref{app:human_validation}. LLMs also assisted with language polishing and editorial organization of the manuscript. All technical content, experimental design, and scientific claims are the work of the authors.

\begin{ack}
We thank Huihui Ling and Haoxiang Zhang for their careful annotation in the human validation studies. This work was carried out during the first author's internship at Tongyi Lab, Alibaba Group. The authors declare no competing interests.
\end{ack}

\bibliography{references}
\bibliographystyle{plainnat}

\newpage
\appendix

\newpage
\appendix

\section{Problem Definition}
\label{app:problem}

We formalize the evaluation pipeline as a two-phase protocol operating over multi-session dialogues with interleaved multimodal media.

\paragraph{Notation.}
Let $\mathcal{C} = \{S_1, S_2, \ldots, S_N\}$ denote the conversation history of a character, where each session $S_i = \{(u_1, a_1), \ldots, (u_{T_i}, a_{T_i})\}$ consists of user--assistant turn pairs. A subset of turns contain multimodal media attachments $m \in \mathcal{M}$, where each $m$ is an image, audio clip, or video file. The full media set across all sessions is $\mathcal{M} = \{m_1, m_2, \ldots, m_L\}$.

\paragraph{Memorization Phase.}
A memory system $\Phi$ processes the conversation history incrementally. At each session $S_i$, the system updates its internal memory state:
\begin{equation}
    \mathcal{K}_{i} = \Phi(\mathcal{K}_{i-1}, S_i, \mathcal{M}_i),
\end{equation}
where $\mathcal{K}_i$ denotes the memory state after processing session $i$, and $\mathcal{M}_i \subseteq \mathcal{M}$ is the set of media files appearing in $S_i$. The memorization phase has no access to future queries, so the system must decide what to store without knowing what will be asked.

For text-only methods operating in \emph{caption} mode, each media file $m$ is first converted to a textual description $\text{cap}(m)$ before ingestion. For multimodal methods in \emph{raw} mode, the original media is preserved as embeddings alongside any textual content.

\paragraph{QA Phase.}
Given a query $q$ and the final memory state $\mathcal{K}_N$, the system first retrieves a relevant subset:
\begin{equation}
    \mathcal{R}_q = \text{Retrieve}(\mathcal{K}_N, q, K),
\end{equation}
where $K$ is the number of retrieved entries. The system then generates an answer:
\begin{equation}
    \hat{y} = \text{LLM}_{\text{QA}}(q, \mathcal{R}_q).
\end{equation}

For long-context models, the retrieval step is bypassed: the entire conversation $\mathcal{C}$ (with all media) is concatenated and fed directly as input alongside $q$.

\paragraph{Evaluation.}
Each query $q$ is paired with a reference answer $y^*$ and an evidence chain $E_q = \{e_1, \ldots, e_H\}$ specifying the dialogue turns and media files required to derive the answer. The QA score is computed by an LLM judge:
\begin{equation}
    \text{Score}(q) = \text{Judge}(\hat{y}, y^*, q) \in \{0, 0.5, 1\}, \qquad q \notin \mathrm{SUM}.
\end{equation}

For SUM-type questions, the dataset provides fixed atomic reference nuggets $\{n_1, \ldots, n_J\}$, and the score is the fraction of nuggets covered (see \S\ref{app:eval_details}).

\paragraph{Capability scope.}
The task types probe complementary uses of historical information: UPD asks for changing entity states, TR for temporal relationships, MH for evidence integration, and SUM for cross-session synthesis. The evidence chains span an average of 3.39 sessions per question. RAG, structured memory, and full-context access are alternative ways to make that history available, so RAG results compare one way of implementing memory with others rather than determine whether a task is a memory task.

\section{Construction Details}
\label{app:construction}

\begin{figure}[t]
\centering
\includegraphics[width=\linewidth]{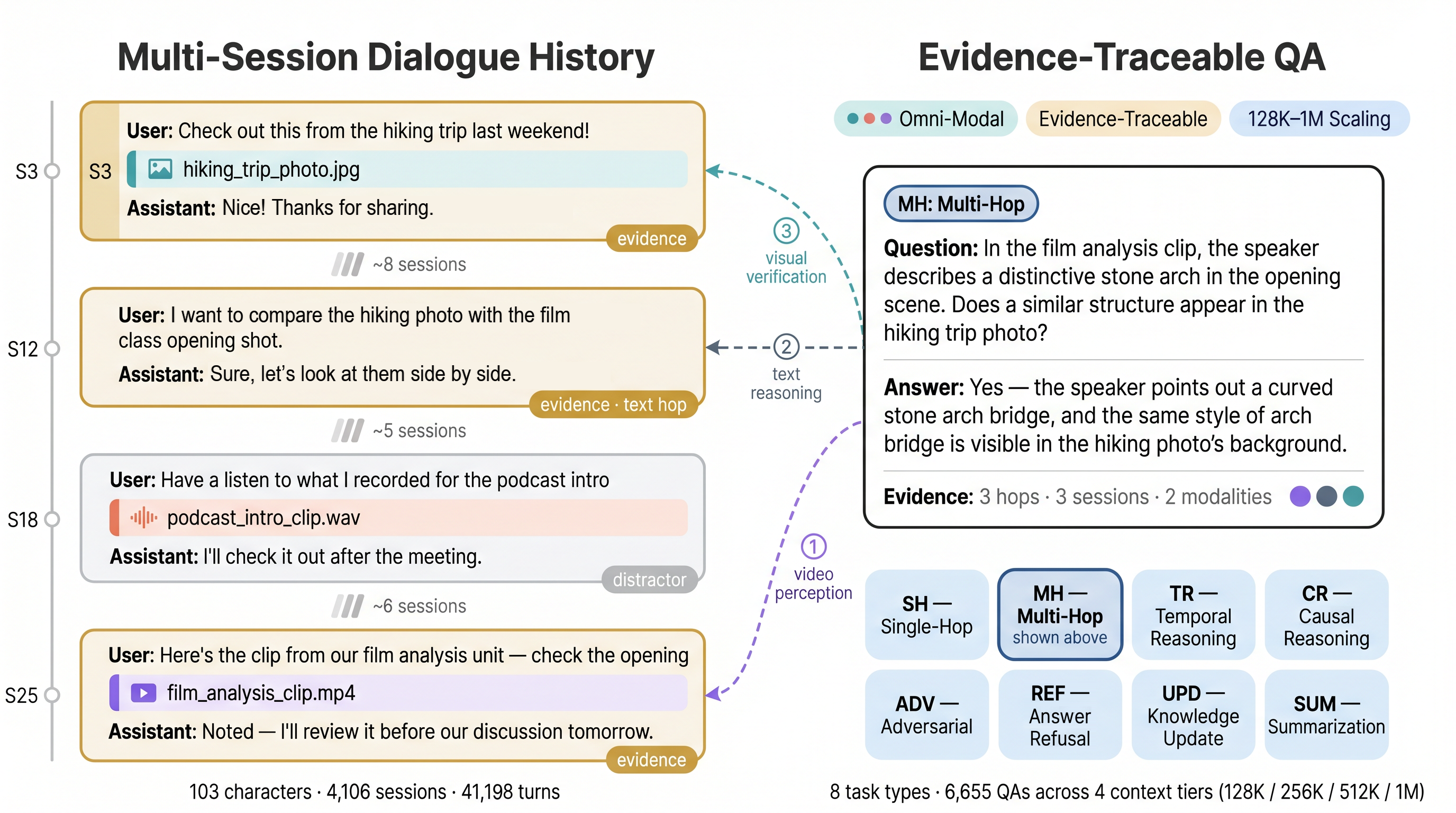}
\caption{OmniMemBench data structure. \emph{Left}: Multi-session dialogue history with interleaved multimodal media (image, audio, video). Evidence sessions are annotated; distractor sessions provide content-isolated padding for context scaling. \emph{Right}: Evidence-traceable QA with multi-hop reasoning chains spanning multiple sessions and modalities. Eight task types range from single-hop extraction to global summarization.}
\label{fig:data_structure}
\end{figure}

\subsection{Data Sources}
\label{app:data_sources}

OmniMemBench draws multimodal content from diverse established research datasets spanning three modalities, totaling 6,987 media files. All sources are widely used benchmarks in their respective communities: MMAU~\citep{sakshi2024mmau} and OmniBench~\citep{li2024omnibench} for audio and cross-modal understanding, MVBench~\citep{li2024mvbench} for video comprehension, Flickr30K~\citep{plummer2015flickr30k} for image captioning, MMMU~\citep{yue2024mmmu}, MMMU-Pro~\citep{yue2024mmmupro}, and ChartQA~\citep{chartqa} for domain-specific visual reasoning, and LibriSpeech~\citep{librispeech}, MusicBench~\citep[released with Mustango]{musicbench}, and ESC-50~\citep{esc50} for speech, music, and environmental sound respectively. Sourcing from established benchmarks rather than web-crawled content provides media files with known content and annotations, which helps construct evidence chains with unambiguous ground-truth answers.

\subsection{Overall Statistics}

Table~\ref{tab:overall_stats} summarizes the benchmark at the 128K tier. The benchmark contains 103 characters, each with an average of 39.9 sessions and 400.0 turns. Multimodal media appears in 20.6\% of turns (audio 9.9\%, image 9.4\%, video 2.4\%). The evidence chains average 3.96 hops (5.0 excluding SUM and SH), spanning an average of 3.39 sessions per question. 95.6\% of non-SUM questions involve exactly one multimodal file.

\begin{table}[ht]
\centering
\caption{Overall benchmark statistics (128K tier, 103 characters). Chain statistics are averaged over all 2,719 QAs.}
\label{tab:overall_stats}
\small
\begin{tabular*}{\linewidth}{@{\extracolsep{\fill}}lr@{}}
\toprule
Statistic & Value \\
\midrule
Characters & 103 \\
Total sessions & 4,106 \\
Avg.\ sessions per character & 39.9 \\
Total turns & 41,198 \\
Avg.\ turns per character & 400.0 \\
Turns with media & 8,487 (20.6\%) \\
Total QAs & 2,719 \\
Avg.\ QAs per character & 26.4 \\
Evidence media files (unique within character) & 2,533 \\
Avg.\ evidence hops per QA & 3.96 \\
Avg.\ evidence sessions per QA & 3.39 \\
Task types & 8 \\
Context tiers & 4 (128K / 256K / 512K / 1M) \\
\bottomrule
\end{tabular*}
\end{table}

\subsection{Type Assignment}
\label{app:type_assignment}

Each non-SUM QA is assigned one of seven task types via an LLM-based fitness scoring procedure; SUM questions are generated separately after the dialogues are complete. For each candidate QA, the generation model reads the captions of its multimodal items and the relations among them, and returns its three best-fitting types, each with a fitness score on a 1--5 scale. We then apply greedy allocation: QAs are sorted by their top fitness score and assigned to the highest-scoring type, subject to per-character quotas of 15\% of the candidates for each of MH, TR, CR, ADV, UPD, and REF and 10\% for SH, with any rounding remainder assigned to MH. If that type has reached its quota, the QA takes its next-best type; if all three have, it takes the type with the most remaining quota. This keeps the task types balanced while following each QA's best-fitting types. Later quality filtering changes the final shares (Table~\ref{tab:task_types}).

The detailed task types are as follows:

\begin{table}[t]
\centering
\caption{Eight task types in OmniMemBench. QA counts are for the 128K tier (103 characters).}
\label{tab:task_types}
\small
\begin{tabular}{llrl}
\toprule
Type & Name & \#QA & Description \\
\midrule
SH & Single-Hop & 315 & Retrieve a single fact directly from memory. \\
MH & Multi-Hop & 547 & Integrate multiple pieces of information to reason. \\
TR & Temporal Reasoning & 375 & Reason over temporal relationships across sessions. \\
CR & Causal Reasoning & 323 & Trace cause-effect chains grounded in memory. \\
ADV & Adversarial & 347 & Identify and reject questions with false premises. \\
REF & Answer Refusal & 165 & Refuse to answer when evidence is insufficient. \\
UPD & Knowledge Update & 340 & Track state changes of entities over time. \\
SUM & Summarization & 307 & Synthesize information across multiple sessions. \\
\bottomrule
\end{tabular}
\end{table}
\subsection{Three-Tier Information Control}
\label{app:anti_leakage}

The anti-leakage mechanism operates on three levels of information for each multimodal file:

\begin{itemize}[leftmargin=*,nosep]
    \item \textbf{Caption}: A detailed perceptual description of the media content (e.g., ``The image shows a curved stone arch bridge over a river, with moss-covered stones and a mountain backdrop''). Captions are used internally during QA generation and dialogue planning but are \emph{never} surfaced in dialogue text.
    \item \textbf{Context Hint}: A non-perceptual contextual cue embedded in dialogue (e.g., ``Check out this photo from the hiking trip last weekend!''). Context hints establish when and why the media was shared but do not reveal what it contains.
    \item \textbf{Question}: Probes perceptual details accessible only from the original media (e.g., ``What type of bridge structure is visible in the hiking trip photo?'').
\end{itemize}

During dialogue generation, the generation model receives the caption for planning purposes but is instructed to produce assistant responses that acknowledge the media without describing its content. The multimodal leakage filter (\S\ref{app:qa_pipeline}) verifies this constraint post-hoc, and Appendix~\ref{app:multimodel_leakage} tests the mechanism with all media removed.

\subsection{SUM Generation}

SUM-type questions are generated in two stages after all dialogues are complete:

\begin{enumerate}[leftmargin=*,nosep]
    \item \textbf{Question generation}: The model reviews the full dialogue history for a character and generates summarization questions that require synthesizing information across multiple sessions.
    \item \textbf{Nugget annotation}: Each reference answer is decomposed into 6--10 atomic semantic units, each representing one independently verifiable claim. These nuggets and their source dialogue rounds are fixed in the dataset before system evaluation and shared across methods. During evaluation, the judge labels each nugget as covered or not covered by the predicted answer; the question score is the fraction covered.
\end{enumerate}

\subsection{Quality Assurance Pipeline}
\label{app:qa_pipeline}

From an initial pool of ${\sim}$5,600 candidate QAs, five audit stages progressively filter to 2,719 (48.6\% retention). Each stage targets a distinct failure mode:

\begin{enumerate}[leftmargin=*,itemsep=4pt]
    \item \textbf{Consistency check}: An LLM receives the question, reference answer, and the full evidence chain, then verifies three conditions: (a)~the evidence chain logically supports the answer, (b)~the reasoning steps are valid and do not require information outside the evidence, and (c)~the question is unambiguous. QAs failing any condition are discarded. This stage primarily removes questions with broken reasoning chains or answers that do not follow from the evidence.

    \item \textbf{Grounding check}: For each evidence hop in the chain, we verify that the corresponding information actually appears in the generated dialogue (not just in the planning metadata). We also check that the reference answer is not directly stated in the dialogue text, which would let a model answer without reasoning over the evidence. This stage catches cases where the dialogue generation model failed to embed a planned evidence hop, or inadvertently leaked the answer.

    \item \textbf{Leakage filter}: A text-only model (with all media files removed) attempts to answer each QA using only the dialogue text. QAs that the text-only model answers correctly are filtered out, as they can be answered without the media. The media-removed check in Table~\ref{tab:leakage_multimodel} measures residual text-only answerability after construction.

    \item \textbf{Uniqueness check}: An LLM checks whether the answer is deterministic given the evidence---i.e., whether there exists a plausible alternative answer that is also consistent with the evidence chain. QAs with ambiguous or multiple valid answers are removed.

    \item \textbf{Manual review}: Two authors conducted two audit rounds covering all 6,655 tier-specific QA instances. Each round combined automated screening with manual inspection of flagged items. Approximately 70 dialogue entries were revised; no QAs were removed at this stage. The preceding automated stages had already filtered 51.4\% of base-tier candidates.
\end{enumerate}

\subsection{Generator Robustness}
\label{app:generator_robustness}

To test whether the results depend on the generator, we rebuilt 10 characters with GPT-5.5 in place of Gemini, producing 397 QAs; the characters were selected before the GPT-generated data were inspected. Table~\ref{tab:generator_check} reports five systems on this subset under GLM-5.2, and Kimi-K2.5 yields the same ordering. As in the main benchmark, the two Gemini long-context systems lead NaiveRAG and AUGUSTUS, and Qwen3.5-Omni-Plus ranks last. Because the RAG and memory baselines also answer with Gemini-3-Flash, any familiarity with Gemini-generated text is shared by these systems as well.

\begin{table}[htbp]
\centering
\small
\caption{Generator robustness on 397 QAs generated with GPT-5.5. Scores use GLM-5.2 and the 0--1 scale.}
\label{tab:generator_check}
\begin{tabular}{@{}llr@{}}
\toprule
System & Approach & Score \\
\midrule
Gemini-3-Flash & Long-context & .5698 \\
Gemini-3.1-Pro & Long-context & .5284 \\
NaiveRAG-caption & RAG & .4666 \\
AUGUSTUS-caption & Structured memory & .4285 \\
Qwen3.5-Omni-Plus & Long-context & .2845 \\
\bottomrule
\end{tabular}
\end{table}

\subsection{Multi-Model Media-Removed Check}
\label{app:multimodel_leakage}

Table~\ref{tab:leakage_multimodel} removes all media from the dialogues and has models from four families answer from the text alone, with GLM-5.2 scoring every row. On Core-5, the mean over SH, MH, TR, CR, and UPD, Gemini-3.1-Pro drops from .754 with media to .122 without, and the four text-only models score .122--.208, so these task types depend substantially on the media.

\begin{table}[htbp]
\centering
\footnotesize
\setlength{\tabcolsep}{3pt}
\caption{Media-removed evaluation under GLM-5.2. Except for the first row, models receive dialogue text without media. Core-5 is the macro-average of SH/MH/TR/CR/UPD.}
\label{tab:leakage_multimodel}
\begin{tabular}{@{}lrrrrrrrrr@{}}
\toprule
Model / input & SH & MH & TR & CR & UPD & REF & ADV & SUM & Core-5 \\
\midrule
Gemini-3.1-Pro / media & .816 & .767 & .753 & .828 & .607 & .552 & .907 & .642 & .754 \\
Gemini-3.1-Pro / text & .034 & .066 & .220 & .237 & .052 & .870 & .538 & .585 & .122 \\
Qwen3.7-Max / text & .052 & .059 & .339 & .388 & .052 & .883 & .639 & .595 & .178 \\
GPT-5.5 / text & .147 & .078 & .378 & .393 & .044 & .916 & .301 & .634 & .208 \\
GLM-5.2 / text & .043 & .045 & .293 & .311 & .038 & .909 & .601 & .605 & .146 \\
\bottomrule
\end{tabular}
\end{table}

REF, ADV, and SUM are left out of Core-5 because text-only answers can earn credit on them without recovering media content: REF rewards refusing when evidence is missing, SUM also draws on the dialogue text, and ADV credits rejecting a false premise.

\subsection{Evidence Placement and Video Coverage}
\label{app:construction_audit}

For the 10,455 clue turns of the 2,412 non-SUM QAs at 128K, the relative position of turn $t$ in an $n$-turn session, $(t-1)/(n-1)$, falls in the first third for 30.7\%, the middle third for 35.7\%, and the last third for 33.6\%. The relative position of the clue session in the character history, $i/(N-1)$ for session $i=0,\ldots,N-1$ (0=earliest, 1=latest), averages .31 for MH, .36 for CR, .45 for TR, and .46 for UPD. Clue placement therefore follows no fixed within-session template, and session positions differ by task type.

The non-SUM video-evidence subset contains 265 QAs. Their evidence chains have a median length of 5 hops, the same as for audio and image, with mean lengths of 4.95, 4.38, and 4.49 hops respectively. In our token accounting, a video item averages 6,881 tokens, compared with 343 for audio and 485 for image, so within the 128K budget more video means fewer sessions.

\paragraph{Artifact availability.}
The dataset and evaluation resources are available through the public releases:\\
Code: \url{https://github.com/jshmshijh/OmniMemBench}.\\
Data: \url{https://huggingface.co/datasets/jshmsjh/OmniMemBench}.
The dataset release includes Croissant metadata and a Responsible AI datasheet. The code release includes the construction pipeline with its prompt templates, the evaluation pipeline, leakage filtering, judge scoring, and baseline implementations.

\subsection{Context Tier Statistics}
\label{app:tier_stats}

Table~\ref{tab:tier_stats} summarizes the statistics across the four context tiers. The 128K tier uses all 103 characters; higher tiers use a 51-character subset and are constructed by injecting content-isolated distractor sessions, progressively reducing evidence density from 57.7\% to 8.9\%.

\begin{table}[ht]
\centering
\caption{Context tier statistics. The 128K tier uses all 103 characters; the 256K--1M tiers use a 51-character subset whose type and modality distributions closely match the full set. Evidence sessions remain fixed across all tiers; higher tiers are filled by injecting content-isolated distractor sessions.}
\label{tab:tier_stats}
\small
\begin{tabular*}{\linewidth}{@{\extracolsep{\fill}}lcccc@{}}
\toprule
Tier & Characters & Avg Tokens & Evidence \% & QAs \\
\midrule
128K & 103 & 138K & 57.7\% & 2,719 \\
256K & 51 & 234K & 36.4\% & 1,312 \\
512K & 51 & 468K & 18.2\% & 1,312 \\
1M & 51 & 959K & 8.9\% & 1,312 \\
\bottomrule
\end{tabular*}
\end{table}

\section{Method Details}
\label{app:methods}

We evaluate nine methods spanning three paradigms. The four methods with native multimodal (MM) support are each tested in caption and raw modes, yielding $9 + 4 = 13$ configurations. Our pipeline implementations build on the Mem-Gallery codebase~\citep{bei2026memgallery}, which extends MemEngine~\citep{zhang2025memengine}. Embedding-based retrieval uses Omni-Embed-Nemotron-3B~\citep{xu2025omni}.

\subsection{Long-Context Baseline}

\begin{itemize}[leftmargin=*,itemsep=4pt]
    \item \textbf{SlidingWindow (32K).} Truncates the conversation history to the most recent 32K tokens. This baseline tests whether a recency-only strategy can cover sufficient evidence. It serves as a reference point for memory systems: a method that scores below it gains less from its memory mechanism than from simply keeping the most recent context.
\end{itemize}

\subsection{RAG Methods}

\begin{itemize}[leftmargin=*,itemsep=4pt]
    \item \textbf{NaiveRAG.} Converts all dialogue turns (with media captioned to text) into text chunks, embeds them, and retrieves the top-$K$ most similar chunks given a query. The retrieved chunks are concatenated as context for the QA backbone. This is the simplest RAG baseline with no multimodal-specific design.

    \item \textbf{MultimodalRAG.} Extends NaiveRAG with a multimodal embedding model that encodes both text and media into a shared vector space. In caption mode, media is converted to text before embedding; in raw mode, original media files are embedded directly alongside text. Retrieval and generation follow the same pipeline as NaiveRAG.

    \item \textbf{UniversalRAG}~\citep{yeo2025universalrag}. A modality- and granularity-aware RAG framework that first predicts the target modality and appropriate granularity for each query, then performs targeted retrieval within the predicted corpus. This routing mechanism aims to reduce modality bias and retrieval noise compared to flat retrieval over a unified index.
\end{itemize}

\subsection{Memory Methods}

\begin{itemize}[leftmargin=*,itemsep=4pt]
    \item \textbf{GenerativeAgents}~\citep{park2023generativeagents}. Inspired by the cognitive architecture of generative agents, this method maintains a stream of observations extracted from dialogue turns, periodically synthesizes higher-level reflections from accumulated observations, and retrieves relevant memories using a combination of recency, importance, and relevance scoring.

    \item \textbf{NeuralGraph}~\citep{fisher2025neuralgraph}. Builds a graph-structured episodic memory where nodes represent conversation segments enriched with modality-specific embeddings and temporal metadata. Retrieval operates via graph traversal rather than flat vector similarity, enabling associative, temporal, and cross-modal recall. Supports both caption and raw modes for multimodal content.

    \item \textbf{AUGUSTUS}~\citep{jain2025augustus}. A cognition-inspired multimodal memory system that organizes information into recall memory (raw conversation history) and hierarchical contextual memory (semantic concepts linked to multimodal context). It abstracts user interactions into semantic tags and retrieves information through concept-driven search, supporting both caption and raw modes.

    \item \textbf{AMem}~\citep{xu2025amem}. An agentic memory system that organizes experiences as structured atomic notes inspired by the Zettelkasten method. Each note contains rich contextual descriptions, and new memories autonomously establish links to related past memories, enabling a dynamically evolving memory network.

    \item \textbf{Mem0}~\citep{chhikara2025mem0}. Extracts structured memory entries from each dialogue session through intensive LLM-based processing. The resulting entries are compact: they keep key facts but drop contextual details. Retrieval operates over these extracted entries.
\end{itemize}

\section{Evaluation Details}
\label{app:eval_details}

\subsection{QA Evaluation}

We use GLM-5.2 at temperature 0 for QA and retrieval judgments. The released evaluation pipeline uses concise reference answers for the seven non-SUM task types and the following three-level scoring rubric (Appendix~\ref{app:prompt_eval}):
\begin{itemize}[leftmargin=*,nosep]
    \item \textbf{Score 1 (Correct)}: The answer is accurate and contains all key information required by the reference.
    \item \textbf{Score 0.5 (Partial)}: The answer captures the main concept but misses important details, or is partially correct.
    \item \textbf{Score 0 (Incorrect)}: The answer contradicts the reference, is factually wrong, or fails to address the question.
\end{itemize}

For SUM-type questions, we use the $J$ fixed reference nuggets provided with each question ($6\leq J\leq10$) $\{n_1, \ldots, n_J\}$, each representing one independently verifiable claim. The judge evaluates each nugget independently for presence in the predicted answer, and the score is $\text{Score}_{\text{SUM}}(q) = \frac{1}{J}\sum_{j=1}^{J} \mathbb{1}[n_j \text{ is covered}]$.

\subsection{Aggregation and Comparison Sets}
\label{app:aggregation}

Overall is a question-level micro-average over all valid responses, including SUM: each question contributes one score irrespective of its task type or number of nuggets. Task columns average within the corresponding type. Missing source responses or invalid judge outputs are excluded from the denominator rather than imputed as zero. The main and track tables use all valid responses in each configuration.

Within-method backbone, retrieval-depth, and context-tier QA comparisons use the intersection of questions with valid scores at every compared setting. A question is matched by character, question identifier, and QA index. Context scaling restricts all tiers, including 128K, to the same 51 characters. Each method has a fixed denominator within a paired comparison; methods may have slightly different denominators. The backbone and context-scaling tables report these counts. The per-type context table instead reports all valid responses at each tier and is identified accordingly.

VL-track scores in Table~\ref{tab:vl_track} use expanded reference answers for non-SUM tasks; SUM uses the same fixed nuggets.

CC/ER average per-question rates over valid retrieval judgments at each setting. Empty retrieved sets or empty clue sets are outside this calculation; for the main retrieval set this excludes 23 Mem0 questions. Consequently, retrieval and QA means need not have identical denominators. Retrieval correlations below identify their observation unit explicitly.

\subsection{Retrieval Quality Evaluation}

We propose two LLM-based metrics that enable cross-paradigm retrieval evaluation---applicable to both RAG methods (which return document chunks) and memory methods (which return compressed/restructured entries) without requiring ID-level alignment with ground-truth evidence.

\paragraph{Clue Coverage (CC).}
For each QA, the evidence chain defines a set of clues $\{c_1, \ldots, c_H\}$ (the atomic information units needed to answer). CC measures whether the retrieved entries collectively cover these clues:
\begin{equation}
    \text{CC}(q) = \frac{1}{H} \sum_{h=1}^{H} \text{Judge}_{\text{cover}}(c_h, \mathcal{R}_q),
\end{equation}
where $\text{Judge}_{\text{cover}}(c_h, \mathcal{R}_q) \in \{0, 1\}$ indicates whether clue $c_h$ is semantically present in the retrieved set $\mathcal{R}_q$, as determined by an LLM judge.

\paragraph{Entry Relevance (ER).}
ER measures the fraction of retrieved entries that are relevant to the query:
\begin{equation}
    \text{ER}(q) = \frac{1}{|\mathcal{R}_q|} \sum_{k=1}^{|\mathcal{R}_q|} \text{Judge}_{\text{rel}}(r_k, q, \{c_1, \ldots, c_H\}),
\end{equation}
where $\text{Judge}_{\text{rel}}(r_k, q, \{c_h\}) \in \{0, 1\}$ indicates whether entry $r_k$ contains information relevant to any clue.

CC evaluates recall (``did we find what we need?'') while ER evaluates precision (``is what we found useful?'').

\subsection{Comparison with ID-Based Retrieval Metrics}
\label{app:metric_validation}

We compute P@$K$ and R@$K$ from the overlap between retrieved entry IDs and annotated evidence IDs. Table~\ref{tab:ir_validation} separates three sets. The six-configuration metric-validation set contains GenerativeAgents, NeuralGraph (caption/raw), SlidingWindow, and UniversalRAG (caption/raw). The seven-configuration main retrieval set contains NaiveRAG, MultimodalRAG (caption/raw), AUGUSTUS (caption/raw), AMem, and Mem0. The third set pools those 13 configurations at $K{=}10$ with six configurations at each of $K{=}5,15,20$, yielding 31 method--depth points.

\begin{table}[t]
\centering
\caption{Spearman correlations with ID-based retrieval metrics. Rows use different configuration sets, and $n$ counts method/configuration means rather than questions.}
\label{tab:ir_validation}
\small
\setlength{\tabcolsep}{5pt}
\begin{tabular*}{\linewidth}{@{\extracolsep{\fill}}lrrr@{}}
\toprule
Set & $n$ & CC--R@$K$ & ER--P@$K$ \\
\midrule
Metric-validation set & 6 & 1.000 & 1.000 \\
Main retrieval set & 7 & .786 & .214 \\
Methods and retrieval depths & 31 & .701 & .653 \\
\bottomrule
\end{tabular*}
\end{table}

CC--R@$K$ and ER--P@$K$ both reach Spearman $\rho=1.000$ on the six-configuration validation set, but are .786 and .214 on the main retrieval set. Across all 31 points they are .701 and .653. At the response-record level, the main retrieval set has 18,973 records with correlations .623 and .582; the metric-validation set has 16,300 records with correlations .952 and .971. Records from different methods can refer to the same question, so the records are not independent samples. Content judgments and ID overlap diverge when memory entries paraphrase or merge evidence, because ID overlap then misses content that is still present.

\subsection{Human Validation and Cross-Judge Comparison}
\label{app:human_validation}

We validate the GLM-5.2 judge against independent human annotation and compare it with five additional judges. QA responses, SUM nuggets, and CC/ER clue and entry judgments are different annotation units, so each study is reported separately.

\paragraph{QA response scoring.}
Two annotators independent of the author team independently scored 650 response items spanning seven non-SUM task types and four systems: Gemini-3.1-Pro, NaiveRAG-caption, AUGUSTUS-caption, and AUGUSTUS-raw. Human--human agreement is $\kappa_w=.889$, with 92.6\% exact agreement and Pearson $r=.913$. Against the human reference, GLM-5.2 reaches $\kappa_w=.902$ and 91.8\% exact agreement on these 650 items; Kimi-K2.5 reaches $\kappa_w=.788$ and 83.4\%. Table~\ref{tab:human_tasks} gives the results by task.

\begin{table}[htbp]
\centering
\small
\setlength{\tabcolsep}{5pt}
\caption{QA validation by task on 650 response items. Human, Kimi, and GLM are mean scores; Agr. is exact agreement with the human reference. GLM denotes GLM-5.2. ALL aggregates the 650 items.}
\label{tab:human_tasks}
\begin{tabular}{@{}lrrrrrr@{}}
\toprule
Task & $n$ & Human & Kimi & Agr. (Kimi) & GLM & Agr. (GLM) \\
\midrule
SH  & 96  & .914 & .922 & 97.9\% & .917 & 96.9\% \\
MH  & 88  & .506 & .477 & 86.4\% & .511 & 93.2\% \\
TR  & 100 & .593 & .535 & 87.0\% & .580 & 98.0\% \\
CR  & 112 & .603 & .621 & 88.4\% & .634 & 92.9\% \\
REF & 92  & .660 & .620 & 87.0\% & .560 & 78.3\% \\
ADV & 94  & .910 & .580 & 61.7\% & .915 & 97.9\% \\
UPD & 68  & .408 & .331 & 70.6\% & .287 & 82.4\% \\
\midrule
ALL & 650 & .666 & .596 & 83.4\% & .645 & 91.8\% \\
\bottomrule
\end{tabular}
\end{table}

\paragraph{Cross-judge comparison.}
Table~\ref{tab:six_judges} compares six judges on the 488 items that all six scored, a subset of the 650 items above. Under human annotation, system scores are Gemini-3.1-Pro .773, NaiveRAG .660, AUGUSTUS-caption .641, and AUGUSTUS-raw .590, so the Gemini system still leads and AUGUSTUS-caption still scores above AUGUSTUS-raw.

\begin{table}[htbp]
\centering
\small
\caption{Agreement with the human reference on the common 488-item subset. GLM-5.2 is the benchmark judge; five additional judges provide a comparison on the same response items.}
\label{tab:six_judges}
\begin{tabular}{@{}lrrr@{}}
\toprule
Judge & Weighted $\kappa$ & Exact agreement & Pearson $r$ \\
\midrule
Kimi-K2.5 & .781 & 86.3\% & .813 \\
GLM-5.2 & .918 & 92.6\% & .953 \\
GPT-5.5 & .900 & 90.8\% & .954 \\
DeepSeek-V4-Pro & .907 & 91.8\% & .945 \\
Claude Opus 4.8 & .905 & 92.2\% & .937 \\
Qwen3.7-Max & .911 & 92.2\% & .948 \\
\bottomrule
\end{tabular}
\end{table}

\paragraph{Clue Coverage and Entry Relevance.}
One non-author annotator evaluated 400 QAs across three methods (NaiveRAG-caption, AUGUSTUS-caption, and AUGUSTUS-raw), producing 6,090 binary clue judgments and 10,133 binary entry judgments. Table~\ref{tab:human_ccer} compares these annotations with GLM-5.2. Human and GLM method orderings agree: for CC, NaiveRAG $>$ AUGUSTUS-caption $>$ AUGUSTUS-raw; for ER, AUGUSTUS-caption $>$ AUGUSTUS-raw $>$ NaiveRAG. This annotator came from the same pool as the QA study, whose human--human agreement ($\kappa_w=.889$) serves as the reference.

\begin{table}[htbp]
\centering
\small
\caption{GLM-5.2 agreement with one independent human annotator for retrieval judgments. Counts refer to clue/entry decisions, not distinct QA pairs.}
\label{tab:human_ccer}
\begin{tabular}{@{}lrrr@{}}
\toprule
Metric & Binary judgments & Cohen's $\kappa$ & Exact agreement \\
\midrule
Clue Coverage & 6,090 & .753 & 87.7\% \\
Entry Relevance & 10,133 & .787 & 90.0\% \\
\bottomrule
\end{tabular}
\end{table}

\paragraph{SUM nugget coverage.}
Two non-author annotators assessed 1,828 nugget decisions from 50 QAs across four systems. Human--human exact agreement is 83.3\%. Among the 1,523 decisions on which both annotators agree, GLM-5.2 agrees with their label on 98.1\%. The human system ordering is Gemini-3.1-Pro $>$ NaiveRAG $>$ AUGUSTUS-caption $>$ AUGUSTUS-raw; GLM preserves this ordering, with the two AUGUSTUS variants nearly tied.

\paragraph{Reproducibility.}
Evaluator runs use temperature 0. Because API-served models can change over time, we report the model identifiers, prompts, and scoring configuration used for every result.

\subsection{Backbone Ablation}
\label{app:backbone}

Table~\ref{tab:backbone} compares a store-backbone change (Qwen3.5-35B-A3B to 122B-A10B, with QA=Flash) and a QA-backbone change (Gemini-3-Flash to Gemini-3.1-Pro, with store=35B). The larger store backbone yields mixed, small changes, while the Pro QA backbone scores lower for every configuration in this comparison.

\begin{table}[t]
\centering
\caption{Backbone comparisons on common valid questions ($n$) within each method. Store comparisons hold QA=Flash fixed; QA comparisons hold store=35B fixed. $\Delta$ is the second score minus the first.}
\label{tab:backbone}
\begin{minipage}[t]{0.49\linewidth}
\centering\footnotesize
\textbf{Store backbone: 35B vs.\ 122B}\\[3pt]
\setlength{\tabcolsep}{2pt}
\begin{tabular*}{\linewidth}{@{\extracolsep{\fill}}lrrrr@{}}
\toprule
Method & $n$ & 35B & 122B & $\Delta$ \\
\midrule
NaiveRAG$^{\mathrm{c}}$ & 2719 & .606 & .605 & $-$.001 \\
MM-RAG$^{\mathrm{c}}$ & 2716 & .604 & .604 & +.000 \\
UniversalRAG$^{\mathrm{c}}$ & 2716 & .528 & .533 & +.005 \\
AUGUSTUS$^{\mathrm{c}}$ & 2711 & .521 & .505 & $-$.016 \\
MM-RAG$^{\mathrm{r}}$ & 2717 & .476 & .477 & +.001 \\
UniversalRAG$^{\mathrm{r}}$ & 2713 & .466 & .472 & +.006 \\
AUGUSTUS$^{\mathrm{r}}$ & 2686 & .464 & .454 & $-$.010 \\
NeuralGraph$^{\mathrm{c}}$ & 2719 & .452 & .451 & $-$.001 \\
NeuralGraph$^{\mathrm{r}}$ & 2717 & .376 & .372 & $-$.004 \\
Mem0$^{\mathrm{c}}$ & 2718 & .273 & .264 & $-$.009 \\
Gen.Agents$^{\mathrm{c}}$ & 2718 & .259 & .259 & +.000 \\
SlidingWin$^{\mathrm{c}}$ & 2719 & .210 & .202 & $-$.007 \\
\bottomrule
\end{tabular*}
\end{minipage}
\hfill
\begin{minipage}[t]{0.49\linewidth}
\centering\footnotesize
\textbf{QA backbone: Flash vs.\ Pro}\\[3pt]
\setlength{\tabcolsep}{2pt}
\begin{tabular*}{\linewidth}{@{\extracolsep{\fill}}lrrrr@{}}
\toprule
Method & $n$ & Flash & Pro & $\Delta$ \\
\midrule
NaiveRAG$^{\mathrm{c}}$ & 2719 & .606 & .579 & $-$.027 \\
MM-RAG$^{\mathrm{c}}$ & 2719 & .603 & .574 & $-$.030 \\
UniversalRAG$^{\mathrm{c}}$ & 2717 & .528 & .507 & $-$.021 \\
AUGUSTUS$^{\mathrm{c}}$ & 2715 & .520 & .471 & $-$.049 \\
MM-RAG$^{\mathrm{r}}$ & 2717 & .476 & .428 & $-$.048 \\
UniversalRAG$^{\mathrm{r}}$ & 2714 & .466 & .457 & $-$.008 \\
AUGUSTUS$^{\mathrm{r}}$ & 2687 & .464 & .408 & $-$.056 \\
NeuralGraph$^{\mathrm{c}}$ & 2719 & .452 & .398 & $-$.054 \\
AMem$^{\mathrm{c}}$ & 2719 & .438 & .358 & $-$.080 \\
NeuralGraph$^{\mathrm{r}}$ & 2719 & .376 & .297 & $-$.080 \\
Mem0$^{\mathrm{c}}$ & 2718 & .273 & .174 & $-$.099 \\
Gen.Agents$^{\mathrm{c}}$ & 2718 & .259 & .179 & $-$.079 \\
SlidingWin$^{\mathrm{c}}$ & 2719 & .210 & .136 & $-$.074 \\
\bottomrule
\end{tabular*}
\end{minipage}
\end{table}

\paragraph{Store backbone.}
On common questions, changes range from $-$.016 to +.006, with no consistent gain from the 122B model. Embedding-based pipelines such as NaiveRAG do not call the store LLM at all, so their small differences reflect run-to-run variation rather than storage. These results concern the tested store procedures and the 35B/122B pair.

\paragraph{QA backbone.}
Holding stored and retrieved content fixed, switching from Gemini-3-Flash to Gemini-3.1-Pro decreases Overall in all 13 configurations, by .008--.099. We use Qwen3.5-35B-A3B + Gemini-3-Flash as the default configuration throughout the pipeline experiments.

\section{Extended Experimental Results}
\label{app:extended}

\subsection{K-Sensitivity Analysis}
\label{app:k_sensitivity}

Table~\ref{tab:k_sensitivity} reports QA score, Clue Coverage (CC), and Entry Relevance (ER) for six configurations supporting all four retrieval depths.

\begin{table}[t]
\centering
\caption{Retrieval-depth comparison. Each triplet is QA / CC / ER. QA uses a fixed question intersection across all four $K$ values; CC/ER use valid retrieval judgments at each $K$.}
\label{tab:k_sensitivity}
\footnotesize
\setlength{\tabcolsep}{1.8pt}
\begin{tabular*}{\linewidth}{@{\extracolsep{\fill}}lrrrrrrrrrrrr@{}}
\toprule
Method & \multicolumn{3}{c}{$K{=}5$} & \multicolumn{3}{c}{$K{=}10$} & \multicolumn{3}{c}{$K{=}15$} & \multicolumn{3}{c}{$K{=}20$} \\
\cmidrule(lr){2-4}\cmidrule(lr){5-7}\cmidrule(lr){8-10}\cmidrule(lr){11-13}
 & QA & CC & ER & QA & CC & ER & QA & CC & ER & QA & CC & ER \\
\midrule
NaiveRAG$^{\mathrm{c}}$ & .506 & .680 & .539 & .606 & .769 & .342 & .643 & .798 & .246 & .663 & .822 & .193 \\
MM-RAG$^{\mathrm{c}}$ & .513 & .685 & .539 & .604 & .767 & .343 & .642 & .801 & .246 & .670 & .824 & .193 \\
AUGUSTUS$^{\mathrm{c}}$ & .480 & .623 & .510 & .521 & .668 & .417 & .525 & .674 & .399 & .537 & .680 & .390 \\
MM-RAG$^{\mathrm{r}}$ & .411 & .631 & .522 & .476 & .718 & .328 & .503 & .754 & .237 & .523 & .774 & .184 \\
AUGUSTUS$^{\mathrm{r}}$ & .413 & .590 & .504 & .464 & .637 & .409 & .470 & .649 & .394 & .474 & .662 & .389 \\
AMem$^{\mathrm{c}}$ & .387 & .586 & .612 & .438 & .701 & .451 & .473 & .746 & .348 & .480 & .774 & .284 \\
\bottomrule
\end{tabular*}
\end{table}

On fixed question intersections, QA improves monotonically for all six configurations. From $K{=}5$ to $20$, NaiveRAG and MultimodalRAG caption each gain about .157, AUGUSTUS caption gains .057, AUGUSTUS raw gains .060, AMem gains .093, and MultimodalRAG raw gains .111. Across the same depth range, CC rises and ER falls, showing that additional retrieved entries can improve coverage despite lower average relevance.

\subsection{Retrieval-QA Correlation}
\label{app:retrieval_corr}

Table~\ref{tab:retrieval_corr} relates retrieval means to QA means at $K{=}10$. Across seven configurations, Spearman correlation is .821 for CC--QA and .036 for ER--QA. Exact two-sided permutation tests over all $7!$ rank permutations give $p=.034$ and $p=.963$, respectively. With only seven configurations, this correlation describes a trend rather than an established effect. High entry relevance also does not mean that the retrieved set contains everything needed to answer, since retrieved entries can be relevant yet incomplete.
\begin{table}[t]
\centering
\caption{Retrieval quality and QA at $K{=}10$ for seven configurations. Spearman $\rho=0.821$ for CC--QA and $0.036$ for ER--QA; these are associations between method means.}
\label{tab:retrieval_corr}
\small
\setlength{\tabcolsep}{4pt}
\begin{tabular*}{\linewidth}{@{\extracolsep{\fill}}lrrr@{}}
\toprule
Method & CC & ER & QA \\
\midrule
NaiveRAG$^{\mathrm{c}}$ & .769 & .342 & .606 \\
MM-RAG$^{\mathrm{c}}$ & .767 & .343 & .603 \\
AUGUSTUS$^{\mathrm{c}}$ & .668 & .417 & .520 \\
MM-RAG$^{\mathrm{r}}$ & .718 & .328 & .476 \\
AUGUSTUS$^{\mathrm{r}}$ & .637 & .409 & .464 \\
AMem$^{\mathrm{c}}$ & .701 & .451 & .438 \\
Mem0$^{\mathrm{c}}$ & .517 & .303 & .273 \\
\bottomrule
\end{tabular*}
\end{table}

\subsection{Per-Modality Accuracy}
\label{app:per_modality}

Table~\ref{tab:per_modality} groups each question by its evidence modalities. For non-SUM questions, modalities come from evidence-chain media entries; for SUM, they are the union of media types in the source rounds of multimodal nuggets. A question enters every applicable media group, so columns overlap. The text-only group consists of seven SUM questions whose nuggets are all textual. All valid questions receive at least one modality assignment.

\begin{table}[t]
\centering
\caption{Evidence-modality scores, including SUM, at 128K. Questions may enter several media columns. Full-coverage counts are text-only 7, audio 1,203, image 1,414, video 322; actual counts vary with valid responses. Overall counts each question once.}
\label{tab:per_modality}
\small
\setlength{\tabcolsep}{4pt}
\begin{tabular*}{\linewidth}{@{\extracolsep{\fill}}lrrrrr@{}}
\toprule
Method & Text-only & Audio & Image & Video & Overall \\
\midrule
NaiveRAG$^{\mathrm{c}}$ & .591 & .580 & .601 & .473 & .606 \\
MM-RAG$^{\mathrm{c}}$ & .592 & .575 & .596 & .496 & .603 \\
UniversalRAG$^{\mathrm{c}}$ & .623 & .549 & .502 & .374 & .528 \\
AUGUSTUS$^{\mathrm{c}}$ & .475 & .518 & .486 & .439 & .520 \\
MM-RAG$^{\mathrm{r}}$ & .535 & .455 & .471 & .385 & .476 \\
UniversalRAG$^{\mathrm{r}}$ & .608 & .447 & .471 & .346 & .466 \\
AUGUSTUS$^{\mathrm{r}}$ & .418 & .436 & .453 & .399 & .464 \\
NeuralGraph$^{\mathrm{c}}$ & .402 & .471 & .418 & .323 & .452 \\
AMem$^{\mathrm{c}}$ & .516 & .424 & .426 & .371 & .438 \\
NeuralGraph$^{\mathrm{r}}$ & .418 & .387 & .352 & .288 & .376 \\
Mem0$^{\mathrm{c}}$ & .242 & .297 & .221 & .278 & .273 \\
Gen.Agents$^{\mathrm{c}}$ & .064 & .314 & .208 & .149 & .259 \\
SlidingWin$^{\mathrm{c}}$ & .079 & .244 & .173 & .122 & .210 \\
\bottomrule
\end{tabular*}
\end{table}

\subsection{Context Scaling}
\label{app:scaling}

Table~\ref{tab:scaling_overall} compares six configurations on fixed question intersections across tiers. NaiveRAG and MultimodalRAG caption remain highest, with declines of .017 and .016 from 128K to 1M. AUGUSTUS caption and Mem0 decline by .033 and .022; AUGUSTUS raw changes by only $-$.004. SlidingWindow increases by .030, though its absolute score remains low.

\begin{table}[t]
\centering
\caption{Context scaling on the 51-character subset. Each method uses its common valid question set across all four tiers; $n$ is fixed within its row.}
\label{tab:scaling_overall}
\small
\setlength{\tabcolsep}{4pt}
\begin{tabular*}{\linewidth}{@{\extracolsep{\fill}}lrrrrr@{}}
\toprule
Method & $n$ & 128K & 256K & 512K & 1M \\
\midrule
NaiveRAG$^{\mathrm{c}}$ & 1308 & .606 & .594 & .600 & .589 \\
MM-RAG$^{\mathrm{c}}$ & 1312 & .603 & .586 & .582 & .587 \\
AUGUSTUS$^{\mathrm{c}}$ & 1311 & .508 & .509 & .496 & .475 \\
AUGUSTUS$^{\mathrm{r}}$ & 1277 & .450 & .448 & .444 & .446 \\
Mem0$^{\mathrm{c}}$ & 1310 & .282 & .275 & .271 & .260 \\
SlidingWin$^{\mathrm{c}}$ & 1312 & .213 & .236 & .255 & .242 \\
\bottomrule
\end{tabular*}
\end{table}

Table~\ref{tab:scaling_pertype} gives NaiveRAG-caption task scores using all valid responses at each tier. SH remains high (.979 at 128K and .959 at 1M). MH and REF have the largest endpoint decreases (.581 to .534 and .525 to .463), while other tasks show smaller or non-monotonic changes.

\begin{table}[t]
\centering
\caption{NaiveRAG-caption by task and context tier. These task means use all valid responses at each tier ($n$); Table~\ref{tab:scaling_overall} instead fixes a common set for the Overall comparison.}
\label{tab:scaling_pertype}
\small
\setlength{\tabcolsep}{3pt}
\begin{tabular*}{\linewidth}{@{\extracolsep{\fill}}lrrrrrrrrr@{}}
\toprule
Tier & $n$ & SH & MH & TR & CR & ADV & REF & UPD & SUM \\
\midrule
128K & 1312 & .979 & .581 & .540 & .658 & .933 & .525 & .402 & .210 \\
256K & 1312 & .969 & .552 & .565 & .626 & .915 & .469 & .405 & .205 \\
512K & 1310 & .969 & .556 & .540 & .684 & .921 & .481 & .409 & .213 \\
1M & 1310 & .959 & .534 & .546 & .661 & .939 & .463 & .395 & .203 \\
\bottomrule
\end{tabular*}
\end{table}

\paragraph{Store efficiency scaleup.}
Table~\ref{tab:store_scaleup} reports how memorization time scales with context tier. RAG methods scale about $4\times$ from 128K to 1M. AUGUSTUS grows $14\times$ in caption mode (76$\to$1,069s) and $11\times$ in raw mode (313$\to$3,520s), and Mem0, which runs LLM extraction on every session, grows $5.8\times$ (2,899$\to$16,765s).

\begin{table}[ht]
\centering
\caption{Store time per scenario (seconds) across context tiers, from the context-scaling runs. Times are compared within each row; Table~\ref{tab:efficiency} reports the main 128K runs.}
\label{tab:store_scaleup}
\small
\begin{tabular*}{\linewidth}{@{\extracolsep{\fill}}llrrrr@{}}
\toprule
Method & Mode & 128K & 256K & 512K & 1M \\
\midrule
NaiveRAG & caption & 14 & 19 & 32 & 60 \\
MM-RAG & caption & 16 & 19 & 33 & 61 \\
AUGUSTUS & caption & 76 & 144 & 227 & 1,069 \\
AUGUSTUS & raw & 313 & 370 & 1,454 & 3,520 \\
Mem0 & caption & 2,899 & 4,744 & 7,277 & 16,765 \\
SlidingWindow & caption & 0.2 & 0.4 & 0.6 & 0.7 \\
\bottomrule
\end{tabular*}
\end{table}

\subsection{Efficiency}
\label{app:efficiency}

Table~\ref{tab:efficiency} reports per-scenario wall-clock time at 128K. AMem and Mem0 spend 4,380s and 1,948s on memorization, followed by about 170s for QA. NaiveRAG obtains the highest pipeline score (.606) in 221s per scenario, compared with AMem's 4,548s. Absolute times depend on our implementations and hardware.
\begin{table}[ht]
\centering
\caption{Memorization and QA time per scenario (seconds) in the main 128K runs, sorted by total time (descending). Components and totals are independently rounded.}
\label{tab:efficiency}
\small
\begin{tabular*}{\linewidth}{@{\extracolsep{\fill}}llrrr@{}}
\toprule
Method & Mode & Store (s) & QA (s) & Total (s) \\
\midrule
AMem & caption & 4,380 & 168 & 4,548 \\
Mem0 & caption & 1,948 & 170 & 2,118 \\
AUGUSTUS & raw & 98 & 515 & 613 \\
AUGUSTUS & caption & 107 & 451 & 558 \\
UniversalRAG & raw & 38 & 414 & 452 \\
UniversalRAG & caption & 22 & 321 & 343 \\
MultimodalRAG & raw & 37 & 220 & 256 \\
NeuralGraph & raw & 43 & 208 & 251 \\
NaiveRAG & caption & 13 & 208 & 221 \\
Gen.Agents & caption & 55 & 152 & 206 \\
MultimodalRAG & caption & 13 & 175 & 187 \\
NeuralGraph & caption & 18 & 162 & 180 \\
SlidingWindow & caption & 0.1 & 85 & 85 \\
\bottomrule
\end{tabular*}
\end{table}

\subsection{VL Track Results}
\label{app:vl_track}

The VL track converts audio to captions for models without native audio support. Table~\ref{tab:vl_track} reports results on the 57-character, 1,453-question track. Gemini-3.1-Pro scores .801, Gemini-3-Flash .737, and the other four models .406--.548.

\begin{table}[t]
\centering
\caption{VL track at 128K (57 characters; audio converted to captions), scored by GLM-5.2. Best displayed scores are bold.}
\label{tab:vl_track}
\small
\setlength{\tabcolsep}{2.8pt}
\begin{tabular*}{\linewidth}{@{\extracolsep{\fill}}lrrrrrrrrr@{}}
\toprule
Model & SH & MH & TR & CR & ADV & REF & UPD & SUM & Overall \\
\midrule
Gemini-3.1-Pro & \textbf{.934} & \textbf{.827} & \textbf{.796} & \textbf{.790} & \textbf{.888} & \textbf{.630} & \textbf{.714} & \textbf{.715} & \textbf{.801} \\
Gemini-3-Flash & .908 & .752 & .673 & .781 & .839 & .500 & .684 & .620 & .737 \\
Qwen3.5-Plus & .772 & .465 & .439 & .479 & .660 & .597 & .452 & .629 & .548 \\
Qwen3.5-Omni-Plus & .770 & .284 & .279 & .488 & .539 & .388 & .318 & .552 & .438 \\
Kimi-K2.5 & .751 & .387 & .201 & .448 & .622 & .306 & .285 & .478 & .435 \\
Qwen3-VL-Plus & .689 & .271 & .265 & .475 & .547 & .285 & .294 & .475 & .406 \\
\bottomrule
\end{tabular*}
\end{table}

\section{Case Studies}
\label{app:cases}

We present three case studies illustrating distinct failure modes at each stage of the memory pipeline: memorization (\S\ref{app:case1}), retrieval (\S\ref{app:case2}), and answering (\S\ref{app:case3}).

\begin{figure}[ht]
\centering
\includegraphics[width=0.95\linewidth]{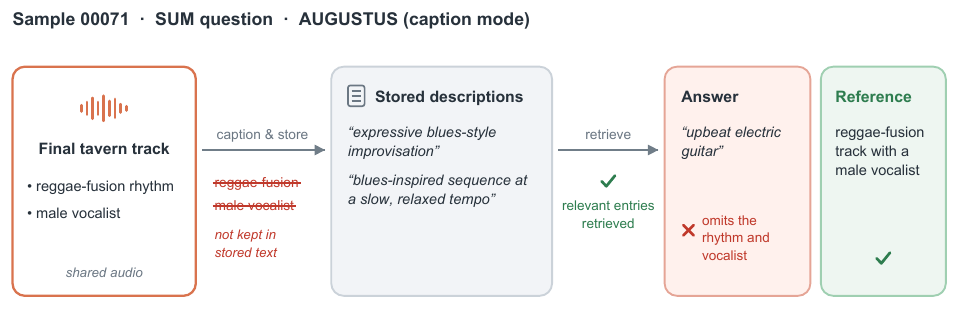}
\caption{A caption-route failure in sample 00071. Retrieved captions describe broad musical properties but omit the final track's reggae-fusion rhythm and male vocalist. The resulting answer does not recover those details.}
\label{fig:case1}
\end{figure}
\subsection{Case 1: Memorization-Time Information Loss (Caption Route)}
\label{app:case1}

\paragraph{Setup.} Sample 00071 asks for the evolution of a game's tavern music direction and the final style selected by the composer. The reference identifies a reggae-fusion piece featuring a male vocalist.

\paragraph{Observed trace.} AUGUSTUS caption stores descriptions including ``expressive blues-style improvisation'' and ``blues-inspired sequence at a slow, relaxed tempo.'' Relevant entries are retrieved, but the generated summary describes the final style as ``upbeat electric guitar,'' omitting the defining rhythm and vocalist.

\paragraph{Analysis.} The stored descriptions in this example do not preserve the details required at query time. Improving retrieval of these same descriptions would not restore the omitted information.

\subsection{Case 2: Retrieval-Time Alignment Failure (Raw Route)}
\label{app:case2}

\begin{figure}[ht]
\centering
\includegraphics[width=0.95\linewidth]{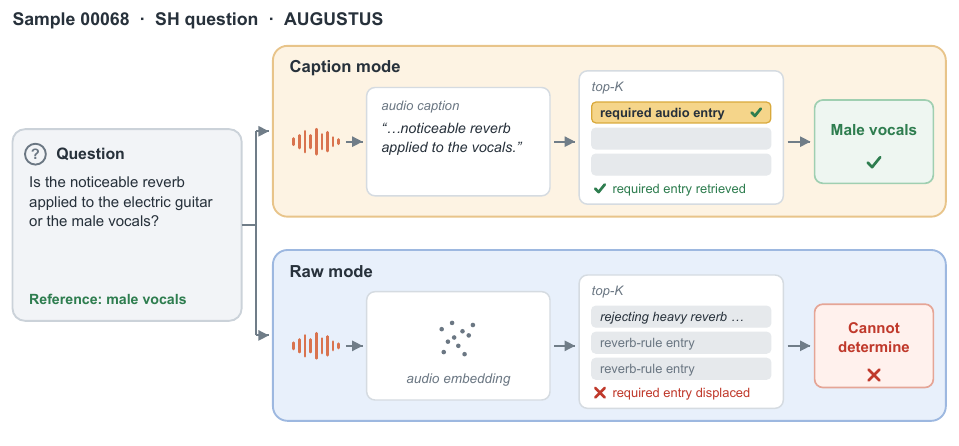}
\caption{Retrieval mismatch in sample 00068. Caption retrieval includes the required track; raw-mode retrieval instead selects entries discussing reverb rules and excludes the required entry.}
\label{fig:case2}
\end{figure}

\paragraph{Setup.} Sample 00068 asks whether noticeable reverb in a groove-oriented track is applied to the electric guitar or the male vocals. The reference answer is ``male vocals.''

\paragraph{Observed trace.} AUGUSTUS caption retrieves the relevant audio description within the top-$K$ and answers correctly. In raw mode, topically related entries about reverb rules, such as rejecting heavy reverb, displace the required entry. The system responds that it cannot determine the answer.

\paragraph{Analysis.} This trace shows how entries on a similar topic can crowd out the specific evidence a question needs, in line with the lower raw-mode scores in Table~\ref{tab:main_results}.

\subsection{Case 3: Visual Perception Failure after Retrieval}
\label{app:case3}

\begin{figure}[ht]
\centering
\includegraphics[width=0.85\linewidth]{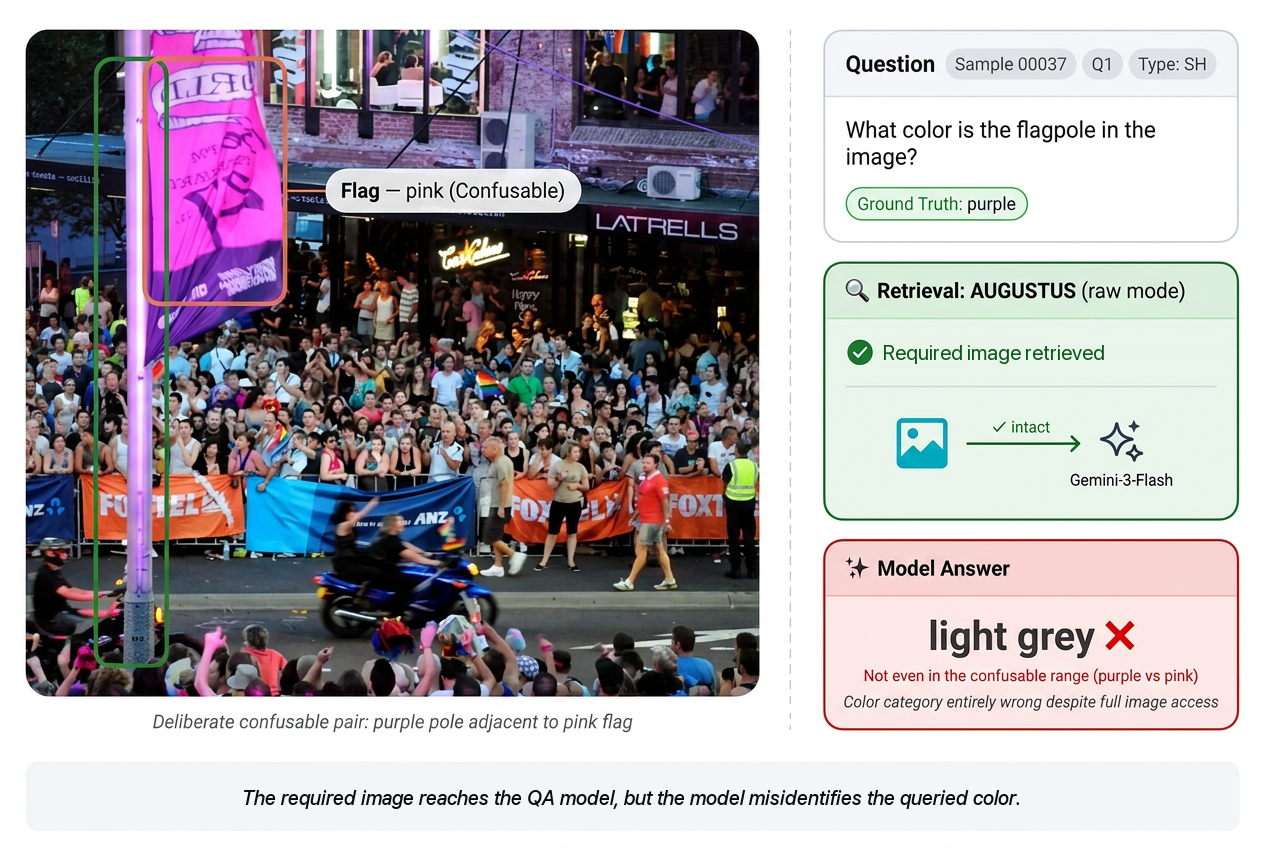}
\caption{Visual perception failure in sample 00037. The required image reaches the QA model, which describes the flagpole as grey; the reference color is purple.}
\label{fig:case3}
\end{figure}

\paragraph{Observed trace.} AUGUSTUS raw retrieves the required image and passes it to Gemini-3-Flash. The image contains a purple flagpole adjacent to a pink flag. Asked for the flagpole's color, the model describes it as grey.

\paragraph{Analysis.} Here, access to the original media is not enough: the QA model misidentifies the queried visual attribute. Together, the three cases show information omitted at storage, a mismatch at retrieval, and a perception error after retrieval.

\newtcolorbox{promptbox}[1][]{
  colback=orange!3, colframe=orange!30,
  coltitle=black,
  fonttitle=\bfseries\small,
  title={#1},
  breakable,
  enhanced,
  left=6pt, right=6pt, top=4pt, bottom=4pt,
  fontupper=\small\ttfamily,
  before upper={\setlength{\parskip}{4pt}\setlength{\parindent}{0pt}},
}

\newpage

\section{Prompt Templates}
\label{app:prompts}

We provide key prompt templates used in the benchmark construction and evaluation pipeline. Placeholder variables are denoted by \texttt{\{variable\}}. Construction prompts are shown in condensed form; the complete templates are included in the code release.

\subsection{Dialogue Generation Prompt}
\label{app:prompt_dialogue}

The following prompt is used in Step~4 of the construction pipeline to generate multi-turn dialogues that naturally embed evidence while preventing information leakage.

\begin{promptbox}[Dialogue Generation Prompt]
You are generating a multi-session dialogue for a multimodal memory benchmark.

The dialogue should:
- Feel like a real person chatting with an AI assistant
- Naturally embed all required evidence
- Follow the phase structure but expand into concrete rounds

\medskip
\textbf{[Meta-Level Understanding --- Read This First]}

The dialogue you generate will become a TEST INPUT for a multimodal memory benchmark. Here is how the benchmark works:

1. A tested model receives ALL session dialogues (including every word from both the user and the assistant) as its memory context.

2. It also receives the actual multimodal files (images, audio, video) embedded in those dialogues.

3. It is then asked a question that requires cross-session reasoning and/or media perception to answer.

This means: the dialogue text is the exam paper, not a real conversation. Every word the user says, every word the assistant says --- all of it is read by the tested model. If the answer to a question can be found in the dialogue text without perceiving the media or without reasoning across sessions, the question becomes trivial and the benchmark is broken.

There are two categories of information that must be protected:

\textbf{Category A --- Multimodal perceptual content:} What a media file depicts, sounds like, or shows. This information must ONLY be obtainable by perceiving the actual media. The dialogue text must not describe, summarize, characterize, or react to the specific content of any media.

\textbf{Category B --- Final assembled answers:} Many questions require collecting evidence fragments from multiple sessions and/or combining text evidence with media perception, then reasoning to reach a conclusion. This final conclusion must NOT appear anywhere in the dialogue. Individual evidence fragments scattered across sessions are fine --- that is the whole point. But no character (user or assistant) should state the assembled result.

Analogy: The exam paper can contain formulas, data tables, and images (evidence fragments). It must NOT contain worked solutions or final answers.

\textbf{Critical distinction:} Text evidence carrying answer-relevant information IS allowed --- it is deliberately planted across sessions for the tested model to find and assemble. What is NOT allowed is anyone in the dialogue performing the assembly and stating the conclusion.

\medskip
\textbf{[Previous Sessions]}\\
\{context\}

\textbf{[Persona]}\\
\{persona\}

\textbf{[Current Session]}\\
\{session\_info\}

\textbf{[Dialogue Plan]}\\
\{dialogue\_plan\}

CRITICAL CONSTRAINT: You MUST use ONLY the evidence pieces listed in the dialogue plan above. Each evidence piece has a specific evidence\_id (e.g., ``Q1\_hop1'', ``Q2\_hop3''). When generating multimodal mm items, the evidence\_id field MUST exactly match an evidence\_id from the dialogue plan. DO NOT create, invent, or hallucinate any additional evidence or evidence\_ids beyond what is provided in the plan.

\medskip
\textbf{[Core Concept: Informed Perception Model]}

You have access to the caption (actual content description) for each multimodal evidence piece, plus the question it serves.

For each multimodal evidence piece you have two layers:

1. \textbf{caption}: Actual content description. Use internally for coherent, non-contradictory planning --- but NEVER expose in output.

2. \textbf{question}: The benchmark question this evidence serves. Tells you exactly what information must NOT be inferable from dialogue text alone.

The precision test: For each multimodal evidence piece: ``If someone reads my entire plan output but never sees/hears the actual media, could they answer the associated question?'' If yes $\rightarrow$ leakage. If no $\rightarrow$ correct.

Determining safe output granularity: Read the caption and question together. Identify every term in the caption that would help answer the question --- those are forbidden in your output. Your output should reference the media at the most generic level that does not leak answer-critical information.

Example: Caption says ``a red acoustic guitar leaning against a wooden chair'', question asks ``What color is the instrument?'':
\begin{itemize}[nosep,leftmargin=*]
\item[\checkmark] ``a hobby-related item'' (generic, does not help answer)
\item[$\times$] ``guitar'', ``red'', ``instrument'', ``wooden chair''
\end{itemize}

\medskip
\textbf{[Anti-Leakage Rules]}

\textbf{Rule 1 --- No media description (Category A):} Check each phrase against the question. If it helps answer $\rightarrow$ remove.

\textbf{Rule 2 --- No final-answer assembly (Category B):} No character assembles scattered evidence into a conclusion.

\medskip
\textbf{[Output Format]}\\
Return a JSON object with fields: session\_id, date, dialogues (array of rounds with user/assistant/mm fields), and summary.
\end{promptbox}

\subsection{QA Generation (Rewrite) Prompt}
\label{app:prompt_qa}

We generate seven QA task types (SH, MH, TR, CR, ADV, REF, UPD), each with a type-specific rewrite prompt. Below we show the Adversarial (ADV) type as a representative example; other types share the same structural template with type-specific rules and subtypes.

\begin{promptbox}[QA Rewrite Prompt (ADV Type)]
You are rewriting multimodal QA data into memory benchmark tasks.

\textbf{Task type: ADV (Adversarial --- False Premise Detection)}

Given an original multimodal QA, produce 5 rewrites. Each rewrite must:
- Present a statement containing a subtle factual error and ask whether it conflicts with dialogue memory
- Require cross-session evidence gathering and multimodal perception to verify
- Be answerable ONLY with the full evidence chain (not from any single hop)

\textbf{Subtypes} (use all across 5 rewrites; no single subtype for all 5):
- Fabricated fact: Insert a false detail that is close to truth (near-miss)
- Misattribution: Attribute correct information to the wrong session/context
- Temporal displacement: Shift timing of events
- Version conflict: Reference outdated information as current

\medskip
\textbf{[Core Rules]}

\textbf{E1. Evidence chain}: $\geq$4 hops per rewrite (recommend 6). Hop counts vary across the 5 rewrites.

\textbf{E2. Temporal safety}: Every text hop's session timestamp $\leq$ multimodal hop's session timestamp. Check each multimodal hop independently.

\textbf{E3. Multimodal irreplaceability}: At least one key fact comes only from multimodal content. No text hop restates it.

\textbf{E4. Hop necessity}: Every hop must be load-bearing. In reasoning, state: ``If hop N removed $\rightarrow$ [specific wrong outcome].''

\textbf{E5. Evidence naturalism}: $\leq$1 explicit text hop per rewrite. Others must embed info as incidental remarks during other discussions. Tag each in reasoning.

\textbf{E6. Session grounding}: Each hop must include session\_id and timestamp matching available sessions.

\textbf{E7. Independent multimodal hops}: Each multimodal file is a separate hop with its own mm\_id (string). Do NOT combine multiple files into a single hop.

\medskip
\textbf{[Reference Style]}

\textbf{R1. Event context only}: Reference multimodal content by occasion, timing, or purpose --- never by perceptual descriptions.\\
\checkmark~``the desk photo I shared during the workspace audit''\\
$\times$~``the photo showing a keyboard and succulent''

\textbf{R2. Unambiguous identification}: References must uniquely identify the target session among all available sessions.

\medskip
\textbf{[Input]}

Persona: \{persona\_info\}\\
Available Sessions: \{available\_sessions\}\\
Question Session ID: \{question\_session\_id\}\\
Original QA --- ID: \{question\_id\}, Question: \{original\_question\}, Answer: \{original\_answer\}, Modalities: \{modalities\}, Multimodal Context: \{caption\}

\medskip
\textbf{[Output Format]}

Return a JSON object with fields: original\_question\_id, original\_question, original\_answer, modalities, and rewrites (array of 5 objects each containing rewrite\_id, rewritten\_question, short\_answer, full\_answer, task\_type, task\_subtype, evidence\_chain, and reasoning).

Each evidence\_chain hop contains: type (text/image/video/audio), session\_id, timestamp, evidence, and mm\_id (for multimodal hops).

Reasoning must include: (a) claim verification with counter-intuitive identification, (b) hop necessity with removal test, (c) multimodal irreplaceability check, (d) naturalism tags per text hop, (e) temporal safety check.
\end{promptbox}

\subsection{Evaluation Prompts}
\label{app:prompt_eval}

\subsubsection{QA Judge Prompt}

We use a unified three-tier scoring system across all task types (SH, MH, TR, CR, ADV, REF, UPD).

\begin{promptbox}[QA Evaluation (LLM-as-Judge) Prompt]
You are an impartial judge evaluating the memory capabilities of an AI assistant with the question-answering task. Your task is to compare the Assistant's Answer against the Ground Truth and assign a score of 0, 0.5, or 1.

\medskip
\textbf{Scoring Rubric}

\textbf{Score 0 (Incorrect / Miss):}
\begin{itemize}[nosep,leftmargin=*]
\item The answer contradicts the Ground Truth or provides factually wrong information.
\item For Yes/No questions: The answer has the wrong polarity (e.g., says ``Yes'' when Ground Truth is ``No'').
\item For Open-ended questions: The answer provides hallucinated content or misses the core entity entirely.
\item For Adversarial questions (questions with false premises): The assistant fails to identify the false premise and provides a fabricated answer.
\item The assistant fails to provide the required information.
\end{itemize}

\textbf{Score 0.5 (Partial):}
\begin{itemize}[nosep,leftmargin=*]
\item The answer captures the main entity or concept correctly but is incomplete, vague, or lacks key supporting details present in the Ground Truth.
\item For Yes/No questions: The polarity is correct but the reasoning is flawed, or the assistant expresses significant uncertainty.
\item For Adversarial questions: The assistant expresses doubt about the premise but still provides an answer without clearly rejecting it.
\item The answer contains correct core information mixed with minor hallucinations or unnecessary filler.
\end{itemize}

\textbf{Score 1 (Correct):}
\begin{itemize}[nosep,leftmargin=*]
\item The answer is accurate, confident, and covers all core information required by the Ground Truth without hallucinations.
\item For Yes/No questions: The polarity matches perfectly.
\item For Adversarial questions: The assistant clearly identifies the false premise and refuses to provide a fabricated answer.
\item Minor wording differences or additional but non-contradictory details are acceptable.
\end{itemize}

\medskip
\textbf{Input Data}

Question: \{question\}\\
Ground Truth: \{ground\_truth\}\\
Assistant Answer: \{model\_output\}

\medskip
\textbf{Output Format}

Output strictly in the following JSON format:\\
\{``score'': <0, 0.5, or 1>, ``reasoning'': ``<short explanation>''\}
\end{promptbox}

\subsubsection{Summary (Nugget-Based) Judge Prompt}

For SUM-type questions, the judge receives the fixed atomic reference nuggets stored with the dataset question. The same nuggets are used for every system; the judge labels their coverage without redefining them.

\begin{promptbox}[Summary Evaluation (Nugget-Based) Prompt]
You are evaluating a model's summary answer for a multi-session dialogue memory benchmark.

Question: \{question\}

Model Output:\\
\{model\_output\}

Below are atomic nuggets that a correct summary should mention. For each nugget, judge whether the model output covers it (1) or not (0). A nugget is ``covered'' if the model output conveys the same information, even if phrased differently. Partial or vague mentions do NOT count.

Nuggets:\\
\{nuggets\_list\}

\medskip
\textbf{Output Format}

Return a JSON object mapping each nugget\_id to 1 (covered) or 0 (not covered). Output JSON only, no explanation.
\end{promptbox}

\subsubsection{Retrieval Quality (Evidence) Judge Prompt}

This prompt jointly evaluates two retrieval metrics --- Clue Coverage (recall) and Entry Relevance (precision) --- in a single LLM call.

\begin{promptbox}[Retrieval Quality Evaluation Prompt]
You are evaluating the retrieval quality of a memory system. Given a question, its ground-truth evidence clues, and the retrieved memory entries, you need to make two independent judgments:

1. \textbf{Clue Coverage}: For each evidence clue, judge whether ANY retrieved entry semantically covers its core information (true) or not (false).

2. \textbf{Entry Relevance}: For each retrieved entry, judge whether it is relevant to ANY evidence clue (true) or not (false).

These two judgments are independent --- a clue being uncovered does NOT imply the closest retrieved entry is irrelevant, and vice versa. Judge each item on its own merit.

A clue is ``covered'' if a retrieved entry conveys the same core information, even if rephrased or consolidated. A retrieved entry is ``relevant'' if it contains information that substantively overlaps with at least one clue. Vague or tangential overlap does NOT count for either direction.

\medskip
\textbf{Input Data}

Question: \{question\}

Evidence Clues:\\
\{clue\_texts\}

Retrieved Memory Entries:\\
\{retrieved\_texts\}

\medskip
\textbf{Output Format}

Output strictly in the following JSON format:\\
\{``clue\_coverage'': [true/false, ...], ``entry\_relevance'': [true/false, ...], ``reasoning'': ``<short explanation>''\}
\end{promptbox}

\section{Ethics and Risks}
\label{app:ethics}

\subsection{Data Privacy}

OmniMemBench is constructed entirely from synthetic dialogues and publicly available research datasets. All character profiles and conversation histories are generated by LLMs under structured constraints; no real user conversations or interactions are involved. Multimodal media files originate from established research benchmarks (\S\ref{app:data_sources}). The author team reviewed all QA pairs for inaccurate or inappropriate content.

\subsection{Annotation}

Construction auditing was performed by the author team (\S\ref{app:qa_pipeline}). Separate evaluator-validation studies used non-author annotators: two for QA scoring, two for SUM nuggets, and one for CC/ER (Appendix~\ref{app:human_validation}). These studies validate the automated scoring and are separate from the construction audit.

\subsection{Potential Risks}

We identify three potential risks. First, our dialogues are synthetic and cover a limited set of personas, so strong benchmark scores do not guarantee good performance in deployment. Second, the LLMs used in the construction pipeline may introduce distributional biases, such as underrepresenting certain cultural contexts or interaction styles, which carry over into the generated dialogues and QA pairs. Third, the anti-leakage filter (\S\ref{app:qa_pipeline}) systematically removes QAs whose answers are textually inferable, which may bias the benchmark toward questions requiring fine-grained perceptual discrimination at the expense of higher-level multimodal reasoning that happens to leave textual traces. We provide the main construction prompts (Appendix~\ref{app:prompts}; all prompts are in the code release) and data statistics (Appendix~\ref{app:construction}) so that others can assess these biases.

\section{Human Annotation Protocol}
\label{app:annotation_protocol}

\paragraph{Annotators and working arrangements.}
The evaluator-validation annotations were produced by professional data-annotation employees as part of their regular job duties. The annotators were not members of the author team. Compensation followed their regular employment arrangements, and informed consent was obtained. The annotation tasks concern the correctness of model responses and the content of retrieved evidence. The benchmark histories are synthetic and the media come from public research datasets. Annotation sample sizes and agreement statistics are reported in Appendix~\ref{app:human_validation}.

\paragraph{QA decision rules.}
The annotation unit is a model response paired with its question and reference answer. Assign 1 when the response correctly supplies the required information, 0.5 when it captures the main answer but has substantive omissions or partially incorrect details, and 0 when it contradicts the reference, invents the core answer, or fails to supply the requested information. For yes/no questions, incorrect polarity receives 0. For adversarial questions, the response must identify the false premise; expressing doubt while still supplying a fabricated answer does not count as fully correct. Equivalent wording is acceptable. The detailed three-level rubric is given in Appendix~\ref{app:prompt_eval}.

\paragraph{SUM decision rules.}
Use the fixed reference nuggets attached to the question. For each nugget, label 1 if the response conveys the same information and 0 otherwise. Paraphrases count as coverage; vague or partial mentions do not. Evaluate each nugget separately without adding or redefining nuggets. The question score is the fraction of reference nuggets covered.

\paragraph{Retrieval decision rules.}
For CC, examine each reference clue and label whether any retrieved entry conveys its core information. For ER, examine each retrieved entry and label whether its content substantively overlaps with at least one reference clue. These are separate binary decisions: an entry can be relevant without fully covering a clue. Vague topical overlap is insufficient. Clue and entry labels are aggregated into per-question CC and ER as defined in Appendix~\ref{app:eval_details}.

\end{document}